\documentclass[lettersize,journal]{IEEEtran}
\usepackage{amsmath,amsfonts}
\usepackage{array}
\usepackage[caption=false,font=footnotesize,labelfont=rm,textfont=rm]{subfig}
\usepackage{textcomp}
\usepackage{stfloats}
\usepackage{url}
\usepackage{verbatim}
\usepackage{makecell}
\usepackage[table]{xcolor}
\usepackage{graphicx}
\usepackage{cite}
\usepackage{graphicx}%
\usepackage{multirow}%
\usepackage{amsmath,amssymb,amsfonts}%
\usepackage{amsthm}%
\usepackage{mathrsfs}%
\usepackage{xcolor}%
\usepackage{textcomp}%

\usepackage{tabularx}

\usepackage{manyfoot}%
\usepackage{booktabs}%
\usepackage{algpseudocode}%
\usepackage{listings}%
\usepackage{array}
\usepackage{rotating}
\usepackage{lineno}
\usepackage{booktabs}
\begin{document}

\title{Absence is Presence: Understanding Visual Scene Negative Events Under Safety Cognitive Constraint}

\author{
	Zhiyun Jiang,
	Hanyong Wang,
	Binbin Liang,
	Yu Xie,
	Menglong Yang,
	and Wei Li%
	\thanks{Zhiyun Jiang, Hanyong Wang, Binbin Liang, Menglong Yang, and Wei Li are with Sichuan University, Chengdu, China.}%
	\thanks{Yu Xie is with Beijing Institute of Technology, Beijing, China.}%
}


\maketitle

\begin{abstract}
Traditional scene understanding focuses on affirmative information objectively present in images. However, in safety-critical domains, comprehending key information that should exist but is actually absent is vital for risk mitigation. To bridge this gap, we focus on visual scene negative captioning with safety as the cognitive constraint. The core challenge is to convert physical absence into semantic negative events. 
Existing vision-language models (VLMs) struggle with this process because affirmation bias suppresses negative reasoning, while limited mental filling capability and representation bias further hinder the inference of absent information.
 To address these challenges, we propose a negative captioning framework based on counterfactual reconstruction and contrastive decoding (CRCD). 
Inspired by human cognition, CRCD reformulates the task as counterfactual latent change captioning to bypass affirmation bias. It contrasts a synthesized safe expectation with reality to identify semantic omissions. To address limited mental filling, we design a dual-branch counterfactual reconstruction architecture. The amodal completion branch restores defective objects, while the functional association branch infers completely absent safety objects. Concurrently, a multi-condition representation learning mechanism is integrated to mitigate representation bias by projecting universal features onto predefined safety criteria subspaces, thereby capturing information across more dimensions. By decoding feature-level semantic residuals between the reconstructed scene prototype and raw input, CRCD bounds the non-existence search space and activates the decoder's negative logic. Extensive experiments validate the effectiveness of CRCD, establishing a high-performance baseline for this pioneering task.
\end{abstract}

\begin{IEEEkeywords}
Scene Understanding, Negative Event, Cognitive Constraint, Change Captioning.
\end{IEEEkeywords}

\section{Introduction}
\IEEEPARstart{V}{isual} scene understanding aims to identify and detect objects within a scene, infer relationships between objects, and thereby construct a comprehensive semantic framework of the scene. Safety represents a high-level cognitive objective in visual scene understanding\cite{cognitive_goal}. Guided by this principle, the task can be subdivided into multiple research domains including anomaly detection\cite{Anomaly_detection_1}, early warning systems\cite{early_warning}, behavior recognition\cite{anomaly_behavior_1,anomaly_behavior_2}, etc. These applications are widely deployed across scenarios such as autonomous driving\cite{automatic_driving_1,automatic_driving_2}, construction sites, and remote sensing. The accuracy and information richness of scene understanding outcomes are critical prerequisites for ensuring the rationality and reliability of higher-level risk avoidance decisions.

\begin{figure}[h]
	\centering
	\subfloat[]{\includegraphics[width=3.5in]{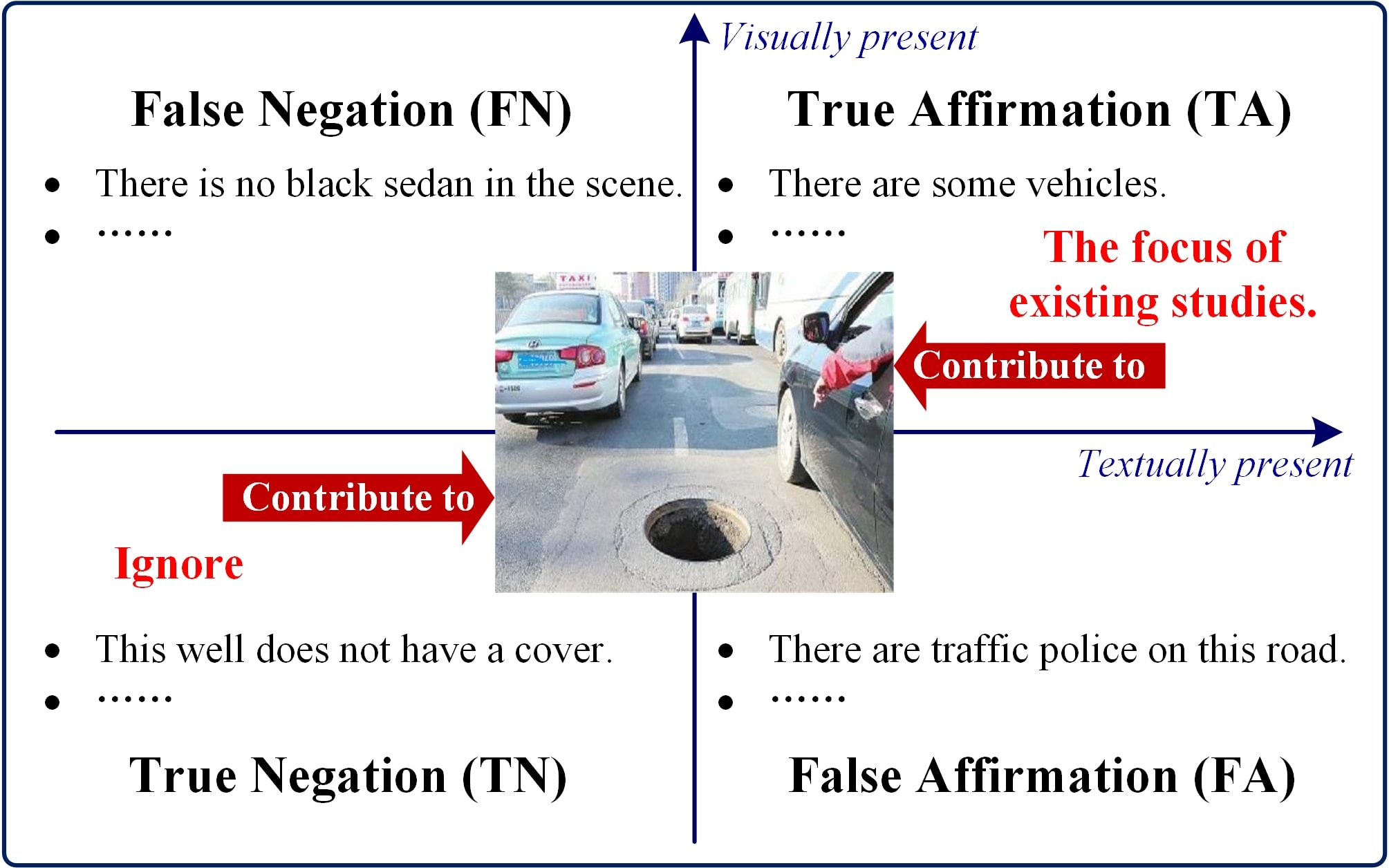}%
		\label{four_results}}
	\hfil
	\subfloat[]{\includegraphics[width=3.5in]{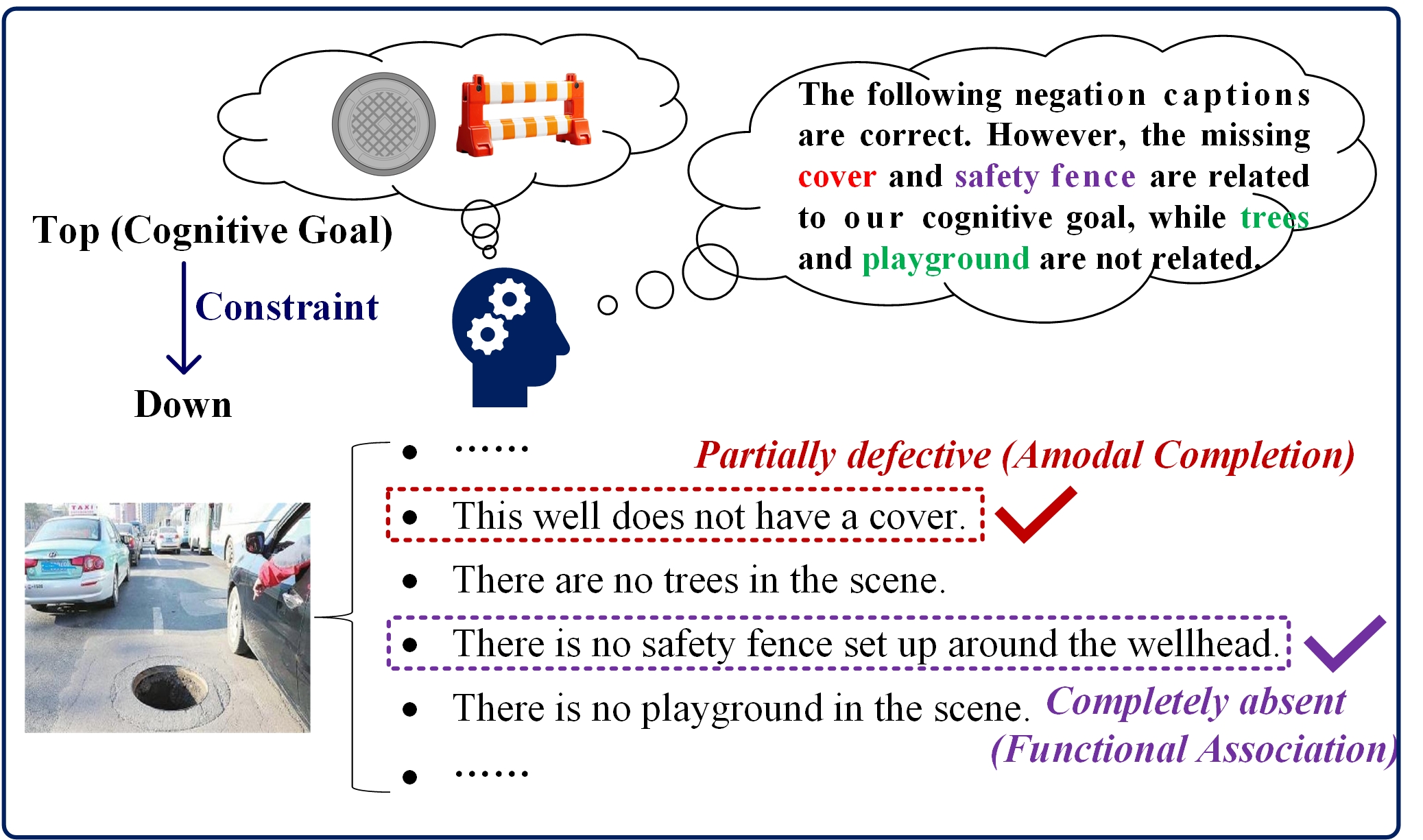}%
		\label{Association}}
	\caption{Our motivation. (a) Four types of scene understanding outcomes. (b) Negative events understanding under the safety cognitive constraint.}
	\label{motivation}
\end{figure}

Translating a visual scene into semantic statements serves as a direct manifestation of visual scene understanding. Theoretically, based on the alignments of visual and textual presence, any such output can be classified into one of four distinct quadrants, as shown in Figure \ref{motivation}\subref{four_results}. Among these, both True Affirmation (TA, where visually present elements are affirmatively described) and True Negation (TN, where visually absent elements are correctly identified through negative semantics) inherently contribute to comprehensive scene cognition. However, existing literature predominantly focuses on optimizing TA to avoid perceptual omissions (i.e., False Negation, FN) and suppress factual hallucinations (i.e., False Affirmation, FA) \cite{hallucination}. Consequently, they largely neglect the exploration of critical, meaningful absent information aligned with high-level cognitive goals (i.e., meaningful TN), thereby restricting current operational boundaries to the raw, visible pixel level. Crucially, weaving these pixel-level absences into high-level scene understanding is fundamentally impossible through conventional affirmative descriptions. Instead, bridging this cognitive gap requires negative semantics (e.g., \textit{no}, \textit{not}, \textit{without}). This shift in the semantic vehicle implies that, under high-level scene cognition, such tactical absence acts not as a perceptual void, but as a distinct form of presence. Functioning as visual dark matter\cite{visual_dark}, these absent factors implicitly delineate the causal and functional boundaries of a scene. As illustrated in Figure \ref{motivation}\subref{Association}, mining such TN is paramount for scene safety understanding, primarily because critical hazards often evade pixel-level detection. Instead, real-world accidents are frequently precipitated by the counterfactual omission of specific objects, relations, or attributes (e.g., a manhole does not have a cover). Consequently, precisely identifying and characterizing this meaningful non-existence provides an essential counter-perspective for risk assessment, while offering interpretable solutions to help decision-makers eliminate potential hazards.

To bridge this literature gap, this study investigates visual negative captioning with safety as the cognitive constraint. Under this formulation, vision-language models (VLMs) are guided to generate negative captions that accurately capture safety risks arising from the absence of critical elements. Compared to conventional image captioning, this task is fundamentally more open-ended and challenging from both computational and philosophical standpoints\cite{negative_caption_difficult}. 
Remarkably, while humans effortlessly achieve such cognitive synthesis\cite{rs_neg}, state-of-the-art VLMs still struggle. We attribute this difficulty to the following three key limitations.

The first is affirmation bias\cite{affirmation_bias}, arising from the dominance of affirmative descriptions in pre-training corpora. The scarcity of negation encourages models to favor positive semantic shortcuts and suppress negative reasoning.
The second is limited mental filling capability\cite{mental_filling}. Unlike humans, who can mentally reconstruct expected visual structures through top-down predictive coding \cite{reconstruction}, existing models struggle both to complete partially defective objects through amodal completion\cite{amodal_completion} and to infer completely absent entities through functional association\cite{causal_inference}.
The third is representation bias\cite{crl}, as standard encoders tend to favor general category and shape features. As a result, they may overlook fine-grained environmental cues, such as lighting, weather, and scene context, that are critical for safety-oriented scene assessment. Consequently, these nested barriers prevent current models from projecting counterfactual safety expectations and mapping semantic discrepancies.

To systematically address these challenges, we propose a negative captioning framework based on counterfactual reconstruction and contrastive decoding (CRCD), which aims to describe safety-related visual non-existence using negative text encompassing absent objects, attributes, and relationships. Specifically, to target the affirmation bias, CRCD reformulates the task into a counterfactual latent change captioning problem. This design is deeply grounded in cognitive psycholinguistics, particularly the two-step simulation hypothesis of negation\cite{neg_cog}. 
The hypothesis posits that human cognition does not process non-existence through direct perception. Instead, it first simulates a counterfactual mental representation of the expected presence and then contrasts it with reality to isolate semantic discrepancies.
 Leveraging this inherent cognitive advantage of contrastive processing, CRCD reconstructs a virtual scene prototype, which acts as an expected scene representation containing information that should exist but is actually absent, and then generates the negative caption through an explicit contrastive decoding process.
This comparative paradigm compresses the non-existence search space from an unbounded domain to a constrained one, thereby reducing task open-endedness and factual hallucinations. The reconstructed prototype also provides a visual anchor for relational semantics, enabling the decoder to capture fine-grained relationships in negative descriptions. Furthermore, to compensate for the model's limited mental filling capability and achieve more effective reconstruction, we design a dual-branch counterfactual reconstruction architecture. 
The amodal completion branch uses cross-modal learning to reconstruct complete semantic prototypes for defective objects. In parallel, the functional association branch serves as a top-down cognitive engine, employing a multi-modal large language model (MLLM) to infer completely absent objects based on perceptual information and the safety cognitive constraint.
 Subsequently, a scene prototype reconstruction module incorporates these heterogeneous results from both branches into corresponding positions within the feature map of the original image via hard and soft injection mechanisms, thereby establishing the virtual scene prototype. Concurrently, we introduce multi-condition representation learning to mitigate universal representation bias. An adaptive router projects visual features onto predefined safety-criterion subspaces, encouraging the model to capture fine-grained safety cues across multiple dimensions. To the best of our knowledge, this is the first method specifically designed to generate negative captions. Overall, the contributions of this paper are summarized as follows:

\begin{itemize}
	\item Inspired by human cognition, we reformulate visual negative captioning as a change captioning task and, accordingly, propose a method framework based on CRCD.
	\item We propose a dual-branch counterfactual reconstruction mechanism to synthesize a virtual scene prototype, providing a robust visual anchor for semantic comparison.
	\item We propose a multi-condition representation learning mechanism to project universal features onto predefined safety criteria subspaces, steering the model to attend to multi-dimensional, fine-grained safety features.
	\item Extensive experiments demonstrate that CRCD significantly outperforms adapted general baselines, establishing a high-performance baseline for this pioneering task.
\end{itemize}

\section{Related Work}

\subsection{Scene Safety Understanding}
Scene safety understanding aims to identify potential hazards, personal protective equipment (PPE)\cite{safety_ppe}, or abnormal behaviors within a scene\cite{safety_abnormal_behavior}, supporting critical applications like autonomous driving\cite{safety_driving1,safety_driving2} and industrial monitoring\cite{safety_construction}. Characterized by its social, prospective, and subjective nature, safety assessment is inherently a high-level cognitive task\cite{cognitive_goal}. Traditional deep learning methods rely on explicit rule-based systems coupled with object detection\cite{safety_ppe}, resulting in fragmented and predefined scene comprehension. In contrast, modern MLLM-based approaches leverage implicit safety knowledge acquired during pre-training to perform advanced reasoning\cite{safety_mllms_1,safety_mllms_2}. However, existing MLLM studies predominantly analyze scene safety based on visible cues, leaving the critical yet invisible negative semantics unexplored.

\subsection{Change Captioning}

Change captioning aims to perceive differences between a reference image and a test image and describe the corresponding changes in natural language. It has been widely studied in remote sensing\cite{change_caption_remote_1,change_caption_remote_2,change_caption_remote_3} and medical imaging\cite{change_caption_medical_1,change_caption_medical_2}. Existing approaches typically learn cross-image correspondence and difference representations before language decoding.
These methods share a contrastive reasoning principle with our task, but conventional change captioning assumes that both visual states are physically available. In contrast, only the actual scene is observed in our setting, while the reference state must be inferred under the safety cognitive constraint. We therefore construct a counterfactual scene prototype and compare it with reality, extending change captioning from observed cross-image differences to semantic discrepancies between reality and expectation.

\subsection{Understanding Negation}
Visual negation understanding concerns the interpretation or generation of language describing non-existence or absence in visual scenes. Early efforts often encoded negation as predefined perceptual categories. For example, PPE monitoring systems distinguish workers with and without helmets as separate detection categories\cite{helmet_detection_1,helmet_detection_2}. Such formulations reduce negation to manually defined mappings rather than contextual reasoning.
Recent studies have increasingly examined negation in VLMs through image--text matching, retrieval, referring expression comprehension, and visual question answering\cite{valse,cc_neg,affirmation_bias,negrefcoco,negation1,NegVQA,medical_neg}. However, most existing tasks determine whether an explicitly specified concept is absent or whether a given negative statement is correct. Negative captioning instead requires the model to autonomously discover meaningful missing information in an open-ended scene. Our task further introduces safety as a cognitive constraint, transforming arbitrary visual absence into goal-relevant negative events.

\section{Methods}

\begin{figure*}[!t]
	\centering
	\includegraphics[width=\linewidth]{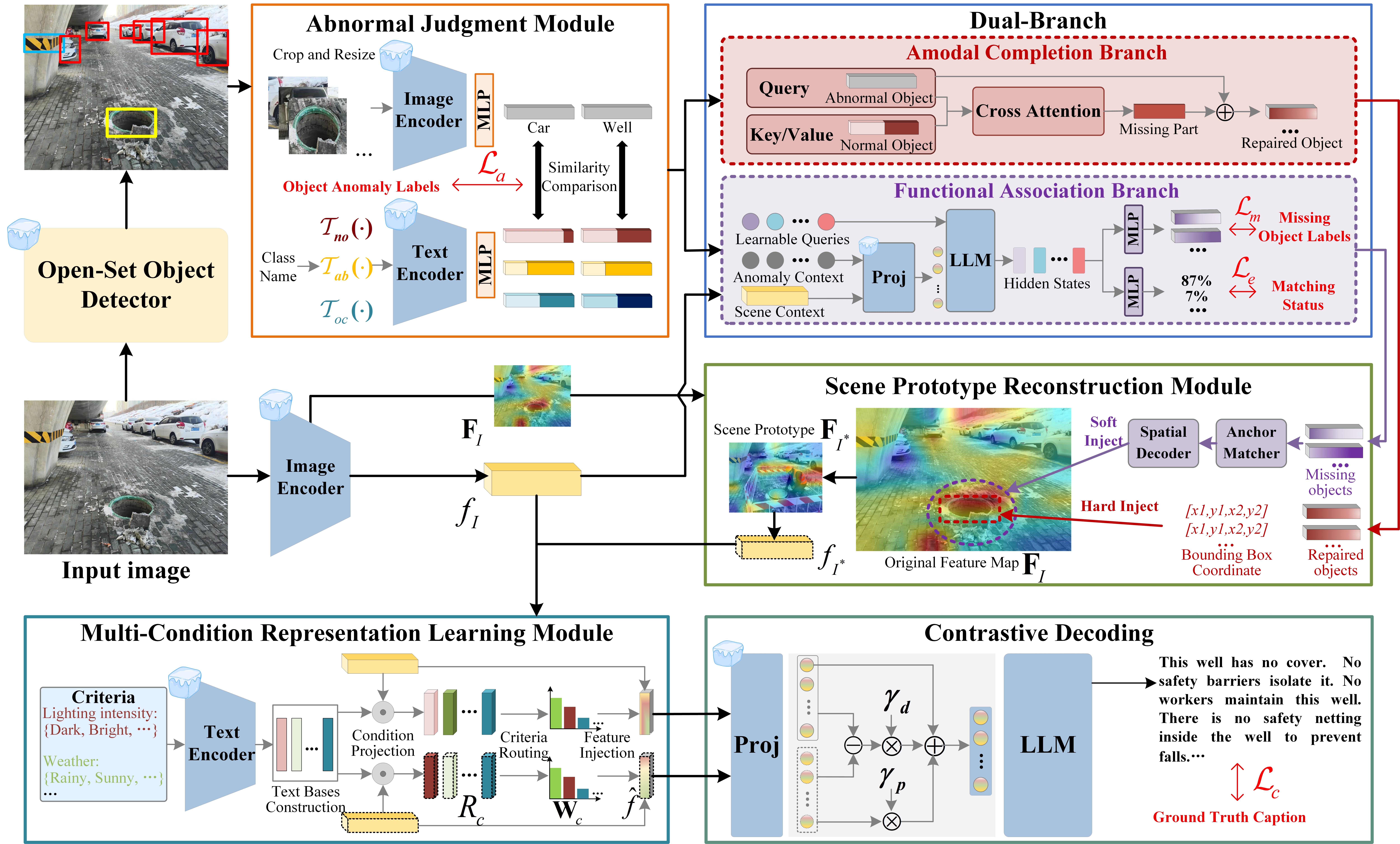}
	\caption{Schematic pipeline of the proposed CRCD framework. Given an input image, the dual-branch network reconstructs a scene prototype via amodal completion and functional association. Simultaneously, the multi-condition representation module projects visual features onto predefined safety criteria. The final negative caption is generated by contrastively decoding the semantic residuals between the reconstructed prototype and the raw input.}
	\label{our_method}
\end{figure*}

\subsection{Overall Architecture}
The overall architecture of our CRCD framework is illustrated in Figure \ref{our_method}. The pipeline unifies visual perception, counterfactual reconstruction, and contrastive language decoding to transform physical absence into semantic negation. Given an input image, the framework first extracts raw visual features and identifies safety-critical anomalies. These cues are funneled into a dual-branch reconstruction network, where an amodal completion branch restores defective objects and a functional association branch infers completely absent safety entities. The outputs of both branches are dynamically integrated with the original features to synthesize a virtual scene prototype representing the fully safe expectation. Concurrently, a multi-condition representation learning module projects the raw features onto predefined safety-criteria subspaces to capture fine-grained, multi-dimensional safety attributes. Finally, by explicitly contrasting the safe scene prototype against the original image, the language decoder isolates feature-level semantic residuals to generate the descriptive negative caption.

\subsection{Existent Object Perception}
Objects constitute the basic elements of scene representation and provide the perceptual basis for inferring absent objects under cognitive constraint. Given an input image $I$, we employ a pre-trained open-set object detector to identify visible objects in the scene. Let $N$ denote the number of detected objects, represented as  $\left\{o_i\right\}_{i=1}^N$. Each object $o_i$ is defined by its category $a_i$ and bounding box $b_i$, i.e., $o_i=(a_i,b_i)$. The detected object regions are then cropped and fed into the abnormal judgment module.

\subsection{Abnormal Judgment Module}
This module aims to leverage the rich prior knowledge of pre-trained VLMs (e.g., CLIP) to determine whether detected objects are defective. For detected object $o_i$, we use the image encoder to extract its visual embedding $v_i \in \mathbb{R}^d$, where $d$ is the embedding dimension. Concurrently, we construct three sets of textual prompts representing distinct states: normal ($\mathcal{T}_{no}$), abnormal ($\mathcal{T}_{ab}$), and occlusion ($\mathcal{T}_{oc}$), and extract their textual embeddings. The safety score $S_{safe}$ and risk score $S_{risk}$ are further defined as:

\begin{equation}
	S_{safe}^i = \text{max}( \langle v_i, \mathcal{T}_{no}(a_i)\rangle, \langle v_i, \mathcal{T}_{oc}(a_i)\rangle )
\end{equation}

\begin{equation}
	S_{risk}^i =\langle v_i, \mathcal{T}_{ab}(a_i) \rangle
\end{equation}
where $\langle \cdot,\cdot \rangle$ denotes cosine similarity. The abnormality probability of $o_i$ is: 
\begin{equation}
	p_i = \frac{\exp(S_{risk}^i / \tau_a)}{ \exp(S_{safe}^i / \tau_a) + \exp(S_{risk}^i / \tau_a) }
\end{equation}
where $\tau_a$ is a learnable temperature parameter. To suppress noise from normal objects, we apply a soft-gating mechanism to the visual feature: $\tilde{v}_i = p_i \cdot v_i$. The gated feature $\tilde{v}_i$ is then fed into the two parallel branches. During training, $\{p_i\}_{i=1}^{N}$ are supervised by object anomaly labels using a binary cross-entropy loss $\mathcal{L}_a$. These labels are directly derived from the negative captions.

\subsection{Dual-Branch}
\subsubsection{Amodal Completion Branch}
To explicitly reconstruct missing parts of damaged or incomplete objects, we introduce an amodal completion branch. This branch restores the fragmented visual feature $\tilde{v}_i$ by retrieving complementary semantic information from a set of ideal text templates $\mathcal{T}_{ideal}$ (e.g., \textit{A complete and intact {$a_i$}}.). We model this completion process as a cross attention mechanism, where the incomplete visual features serve as the query, while the ideal textual features act as the key and value. The extraction of the missing feature $v_{m}^i$ is performed as follows:
\begin{equation}
	v_{m}^i = \text{CA}(query=\tilde{v}_i, key=\mathcal{T}_{ideal}(a_i), value=\mathcal{T}_{ideal}(a_i))
\end{equation}
where $\text{CA}(\cdot)$ refers to the cross attention operation. To adaptively control the strength of feature completion, we further introduce a gating mechanism $\alpha_m^i = \sigma(\text{MLP}(\tilde{v}_i\oplus v_{m}^i))$ where $\sigma$ is the sigmoid function and $\oplus$ denotes concatenation. The final repaired feature $v_{r}^i$ is obtained through residual connections: $v_{r}^i=\tilde{v}_i + \alpha_m^i \odot v_{m}^i$.

\subsubsection{Functional Association Branch}
Although the amodal completion branch can complete explicitly detected incomplete objects, it cannot handle completely missing objects. To address this issue, we introduce the functional association branch. It uses a large language model (LLM) to implicitly infer completely missing objects from global context and anomalous object features, while an expected-existence head estimates whether each inferred object should exist in the counterfactual safe scene.

We define a set of learnable query tokens $\{q_i\}_{i=1}^{N_q}, q_i \in  \mathbb{R}^{d_m}$,  to actively guide the LLM in searching for the features of missing objects. Here, $N_q$ is the predefined number of query tokens, controlling the maximum capacity for representing missing objects, and $d_m$ is the hidden dimension of the LLM. Anomalous objects often imply the expected presence of functionally associated safety entities. To convey object abnormality information to the LLM, we concatenate the gated features of all detected objects to obtain the object abnormality context: $C_{a} =\bigoplus_{i=1}^N \tilde{v}_i$. The LLM receives the original image patch feature $f_{I}$, object abnormality context $C_{a}$, and query tokens $Q=\left[ q_1, ..., q_{N_q} \right]$ as input:

\begin{equation}
	H_a = \text{LLM}_{a}(f_{I}\oplus C_{a}\oplus Q)
\end{equation}
The hidden states corresponding to the query tokens are then projected back into the visual space, generating a set of counterfactual visual vectors $\{v_i^*\}_{i=1}^{N_q}, v_i^*\in\mathbb R^{d_v}$ to represent critical objects entirely absent from the scene, where $d_v$ denotes the channel dimension of the visual encoder's intermediate feature space. Simultaneously, the expected-existence head predicts $\{p_i^*\}_{i=1}^{N_q}$, where $p_i^*$ denotes the confidence that $v_i^*$ corresponds to a valid missing object expected under the safety cognitive constraint. It resolves the fundamental discrepancy between a fixed value of $N_q$ and the actual number of missing objects in real-world scenes, thereby enabling the effective removal of redundant generated counterfactual vectors.

To train this branch more effectively, we use the category labels of missing objects to supervise $\{v_i^*\}_{i=1}^{N_q}$. These labels, denoted by $\{g_i\}_{i=1}^{N_m}$, can be easily obtained from negative captions. Here, $N_m$ represents the number of missing objects. We then use the text encoder to obtain embeddings for these labels, denoted as $\{\hat{g}_i\}_{i=1}^{N_m}$. Concurrently, the generated vectors $\{v_i^*\}_{i=1}^{N_q}$ are projected into the same semantic space, yielding $\{\hat{v}_i\}_{i=1}^{N_q}$. Since the numbers of predicted queries and ground-truth missing objects may differ, we use the Hungarian algorithm\cite{hungarian_algorithm} to obtain the optimal one-to-one matched set $\mathcal{M}_{\pi}$, with $1 - \langle \hat{v}_i, \hat{g}_j \rangle$ as the semantic matching cost. The cardinality of $\mathcal{M}_{\pi}$ is \(K=\min(N_q,N_m)\). Based on this matched set, we define a semantic alignment loss $\mathcal{L}_{m}$ that aligns matched object representations while encouraging diversity among different queries:
\begin{equation}
	\mathcal{L}_{m} = -\frac{1}{K} \!\!\!\sum_{(i,j)\in\mathcal M_\pi}\!\!\!\log \frac{\exp(\langle \hat{v}_i, \hat{g}_j \rangle / \tau_m)}{\sum_{k=1}^{N_q} \exp(\langle \hat{v}_k, \hat{g}_j \rangle / \tau_m)} + \eta \| \hat{\mathbf{V}} \hat{\mathbf{V}}^\top - \mathbf{I} \|_F^2
\end{equation}
where $\tau_m$ is the temperature parameter, $\eta$ is the regularization coefficient, \(\hat{\mathbf{V}} \in \mathbb{R}^{N_q \times d}\) is the matrix representation of $\{\hat{v}_i\}_{i=1}^{N_q}$, \(\mathbf{I}\in \mathbb{R}^{N_q \times N_q}\) is the identity matrix, and \(\| \cdot \|_F\) denotes the Frobenius norm. The first term performs contrastive alignment between matched prediction-label pairs, while the second term encourages diversity among query representations to prevent query collapse. The same matched set $\mathcal{M}_{\pi}$ is also used to supervise the expected-existence head. Queries involved in $\mathcal{M}_{\pi}$ are treated as positives, while the remaining queries are treated as negatives. To mitigate the resulting class imbalance, we employ the focal loss $\mathcal{L}_{e}$:
\begin{equation}
	\mathcal{L}_{e} = \frac{1}{N_q} \sum_{i=1}^{N_q} \begin{cases} 
		-\beta (1 - p_i^*)^\gamma \log(p_i^*), & \exists\, j:\ (i,j)\in\mathcal M_\pi \\ 
		-(1 - \beta) (p_i^*)^\gamma \log(1 - p_i^*), & \text{otherwise} 
	\end{cases}		
\end{equation}
where $\beta \in [0, 1]$ is the weighting factor, and $\gamma \ge 0$ is the focusing parameter.

\subsection{Scene Prototype Reconstruction Module}
This module integrates the dual-branch outputs into the original intermediate feature map $\mathbf{F}_{I}$ to construct a counterfactual latent scene prototype $\mathbf{F}_{I^*}$ that satisfies the safety cognitive constraint. Given $\mathbf{F}_{I} \in \mathbb{R}^{d_v \times H \times W}$, we perform a two-step injection strategy, where \(H\) and \(W\) denote the height and width, respectively.

\subsubsection{Explicit Hard Injection}
We first introduce a linear adaptation layer, which maps the repaired feature $v_r^i$ from the semantic space back to the latent space, yielding the aligned object feature $\hat{f}_{r}^i\in \mathbb{R}^{d_v}$. Subsequently, to avoid boundary artifacts caused by direct feature replacement, we employ a two-dimensional Gaussian distribution for smooth injection. Within the region bounded by the bounding box $b_i$, we establish a local normalized coordinate system $(x_i, y_i) \in [-1, 1] \times [-1, 1]$ and compute the two-dimensional Gaussian weight mask $\mathbf{M}_g^i$:

\begin{equation}
	\mathbf{M}_g^i = \text{clip}\left(\exp\left(-\frac{x_i^2 + y_i^2}{2\phi^2}\right), 0, 1\right)
\end{equation}
where $\text{clip}(\cdot)$ is the truncation function, and $\phi$ is set to control the decay rate. Finally, we broadcast the aligned features $\hat{f}_{r}^i$ in spatial dimensions to obtain $\hat{\mathbf{F}}_{r}^i$. We then iteratively inject the repaired features of all \(N\) detected objects. Let \(\mathbf F_{I^*}^{(0)}=\mathbf F_I\). At the \(i\)-th step, the feature region within \(b_i\) is updated as:
\begin{equation}
\mathbf F_{I^*}^{(i)}(b_i)=\mathbf M_g^i\odot\hat{\mathbf F}_r^i+(\mathbf{1}-\mathbf M_g^i)\odot \mathbf F_{I^*}^{(i-1)}(b_i)
\end{equation}
while features outside \(b_i\) remain unchanged. After all detected objects are processed, we obtain the preliminary counterfactual scene prototype \(\mathbf F'_{I^*}=\mathbf F_{I^*}^{(N)}\).

\subsubsection{Implicit Soft Injection}

While explicit hard injection effectively restores detected defective objects locally, it cannot ground entities predicted by the functional association branch that lack spatial annotations. To address this, we propose an implicit soft injection scheme to dynamically anchor and inject these semantic counterfactual vectors into scene features through an anchor matcher and a spatial decoder.

\begin{figure}[!t]
	\centering
	\includegraphics[width=\linewidth]{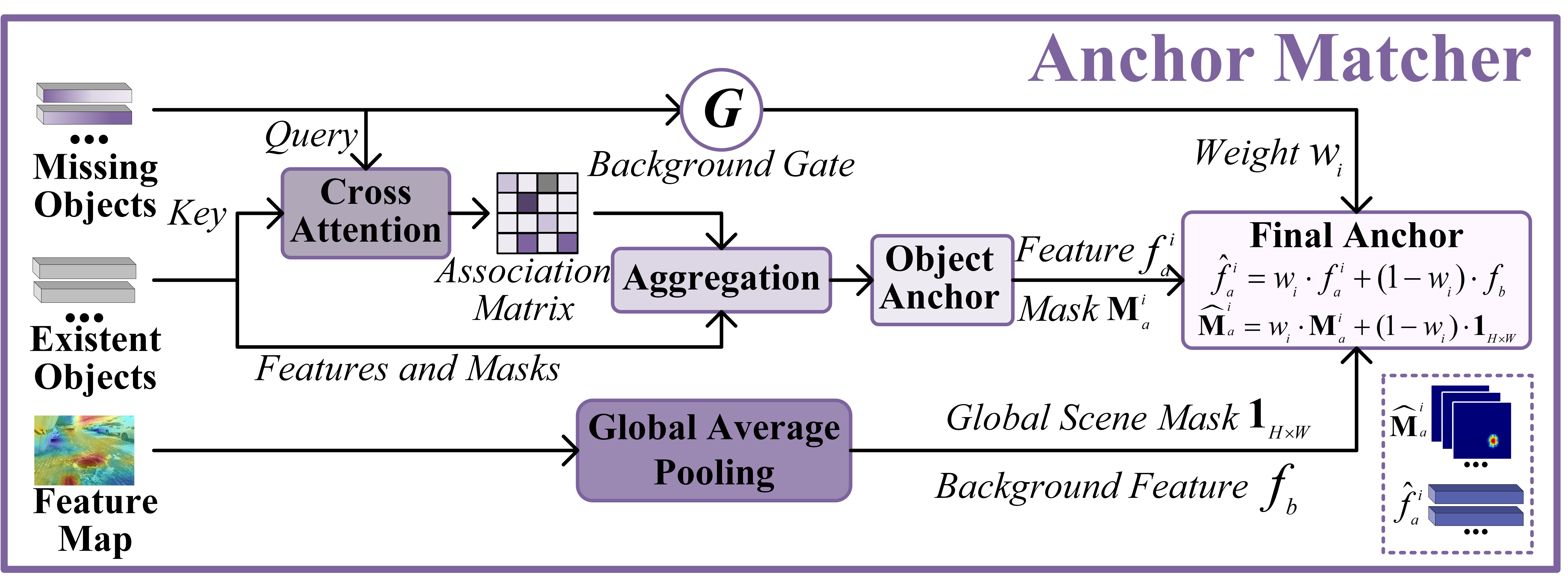}
	\caption{Illustration of the anchor matcher.}
	\label{anchor_matcher}
\end{figure}

These counterfactual objects typically exhibit strong topological associations with existing objects or the overall environment within the scene. The anchor matcher, as illustrated in Figure \ref{anchor_matcher}, computes the association matrix $\mathbf{A}\in\mathbb{R}^{N_q\times N}$ between $\{v_i^*\}_{i=1}^{N_q}$ and the set of existing detected objects $\{o_i\}_{i=1}^{N}$ via a cross-attention mechanism. Based on $\mathbf{A}$, we aggregate the object anchor latent features $\{f_a^i\}_{i=1}^{N_q}, f_a^i \in \mathbb{R}^{d_v}$, and their corresponding spatial masks $\{\mathbf{M}_a^i\}_{i=1}^{N_q}, \mathbf{M}_a^i\in \mathbb{R}^{H\times W}$. To endow the model with adaptive selection capability between local object dependency and global background dependency, we introduce a background gate $w_i = \sigma(\text{MLP}_b(v_i^*))$. The final semantic anchor features $\{\hat{f}_a^i\}_{i=1}^{N_q}$ and anchor masks $\{\hat{\mathbf{M}}_a^i\}_{i=1}^{N_q}$ are dynamically weighted from the object anchor and global background:

\begin{equation}
	\hat{f}_a^i = w_i \cdot f_a^i + (1 - w_i) \cdot f_{b}
\end{equation}

\begin{equation}
	\hat{\mathbf{M}}_a^i = w_i \cdot \mathbf{M}_a^i + (1 - w_i) \cdot \mathbf{1}_{H \times W}
\end{equation}
where $f_{b}$ represents the background latent representation obtained by globally averaging the original feature map $\mathbf{F}_I$. 

\begin{figure}[!t]
	\centering
	\includegraphics[width=\linewidth]{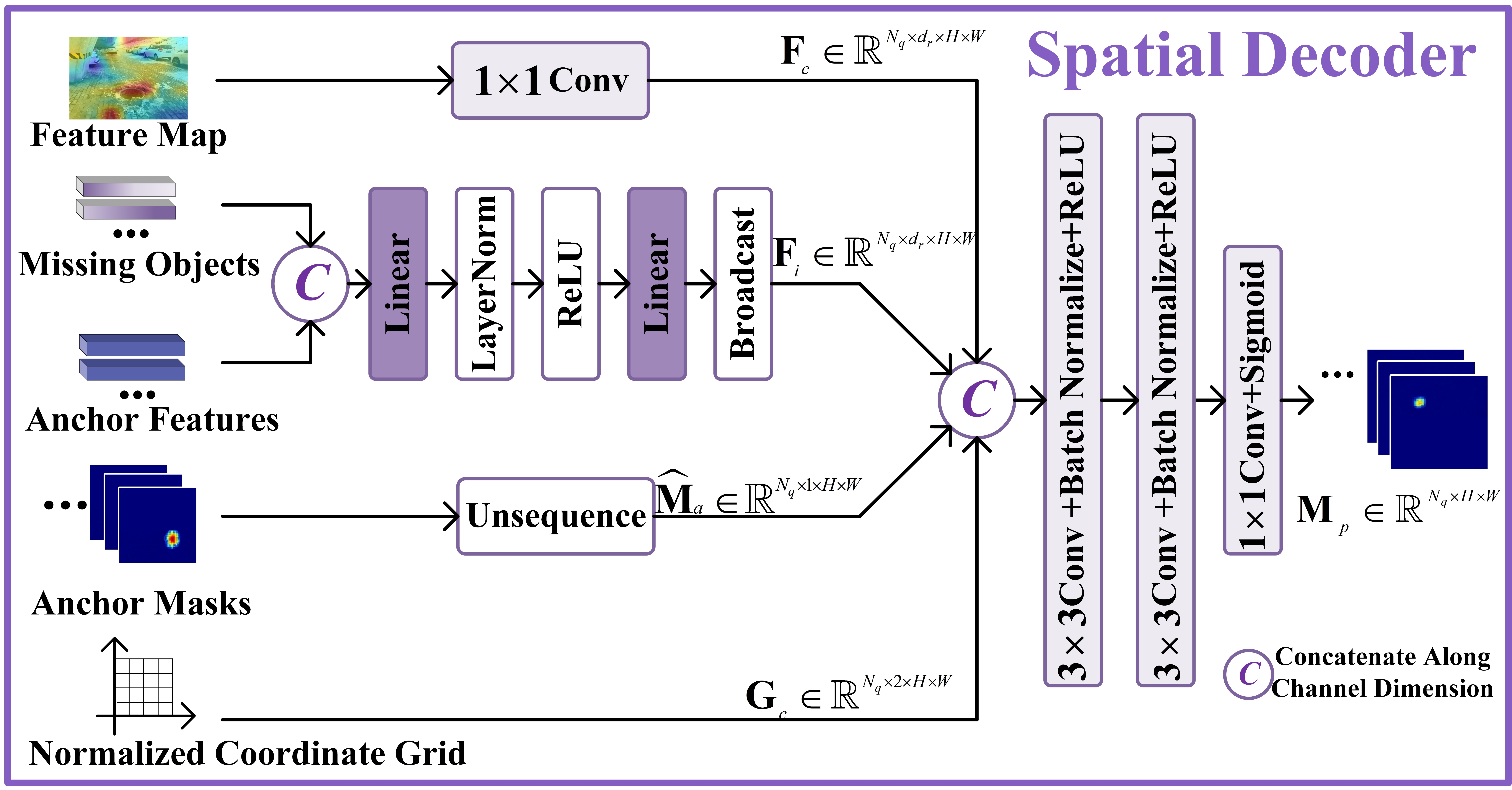}
	\caption{Illustration of the spatial decoder.}
	\label{spatial_decoder}
\end{figure}

After obtaining the semantic anchors, the spatial decoder predicts the precise spatial distribution of the $N_q$ inferred missing objects, as shown in Figure \ref{spatial_decoder}. First, we employ a $1 \times 1$ convolution to perform channel compression on the original intermediate feature map $\mathbf{F}_I$, extracting a compressed scene feature map $\mathbf{F}_{c} \in \mathbb{R}^{N_q \times d_r \times H \times W}$ that embodies scene structure priors, where  $d_r$ is the dimension of the compressed channel. Simultaneously, a linear fusion layer extracts the inferred vectors $\{v_i^*\}_{i=1}^{N_q}$ and their corresponding anchor features $\{\hat{f}_{a}^i\}_{i=1}^{N_q}$, followed by spatial broadcasting after projection to $d_r$ dimensions. This generates the interaction feature map $\mathbf{F}_{i} \in \mathbb{R}^{N_q \times d_r \times H \times W}$ that provides global relational cues.
 We then concatenate $\mathbf{F}_{c}$, $\mathbf{F}_{i}$, the anchor masks $\hat{\mathbf{M}}_{a} \in \mathbb{R}^{N_q \times 1 \times H \times W}$, and normalized coordinate grid $\mathbf{G}_c \in \mathbb{R}^{N_q \times 2 \times H \times W}$, and feed them into a fully convolutional neural network to predict a mask sequence $\mathbf{M}_{p} \in \mathbb{R}^{N_q \times H \times W}$. 
 Each slice $\mathbf{M}_{p}^j$ represents the predicted counterfactual spatial distribution of the \(j\)-th inferred object. Finally, this mask sequence is combined with the existence confidence $\{p_i^*\}_{i=1}^{N_q}$ to guide $\{v_i^*\}_{i=1}^{N_q}$ into the intermediate feature map through weighted accumulation. This achieves precise completion and logical reconstruction of the scene prototype at the feature level, as illustrated below.

\begin{equation}
	\mathbf{F}_{I^*} = \mathbf{F}_{I^*}' + \sum_{j=1}^{N_q} \left(p_j^* \cdot \mathbf{M}_p^j \odot v_j^*\right)
\end{equation}

After feature injection, we feed the scene prototype feature map $\mathbf{F}_{I^*}$ and the original feature map $\mathbf{F}_I$ back into the visual encoder to complete the forward pass through the remaining layers, yielding patch features $f_{I^*}$ and $f_{I}$.

\subsection{Multi-Condition Representation Learning Module}
Scene safety depends on diverse environmental factors that may be underrepresented in generic visual representations. Inspired by conditional representation learning\cite{crl}, we project both the original image and scene prototype features onto criterion-specific subspaces to capture complementary cues associated with predefined safety criteria.

\subsubsection{Text Bases Construction}
Given $N_c$ predefined criteria $\{c_i\}_{i=1}^{N_c}$, we use an LLM to generate descriptive vocabulary for each criterion and construct its semantic basis. The instruction template used is: \texttt{Generate common expressions to describe the [$c_i$] of the scene}. Subsequently, we encode these vocabulary items using the pre-trained text encoder to generate a semantic benchmark matrix $\mathbf{T}_c \in \mathbb{R}^{K_c \times d}$ for criterion $c$. $K_c$ denotes the number of vocabulary items under criterion $c$.

\subsubsection{Condition Projection}
For the extracted visual patch features $f_I,f_{I^*} \in \mathbb{R}^{L\times d}$, where $L$ represents the length of the spatial sequence, we project them into a specific conditional subspace using a generated semantic benchmark matrix, as shown below:

\begin{equation}
	R_c^x = f_x \mathbf{T}_c^\top,\qquad x\in\{I,I^*\}
\end{equation}
Through this operation, we obtain two sets of criterion-specific representations, \(\{ R_{c_i}^{I}\}_{i=1}^{N_c}\) and \(\{ R_{c_i}^{I^*}\}_{i=1}^{N_c}\), for the original image and scene prototype, respectively.

\subsubsection{Criteria Routing}
Given that different regions of an image exhibit varying sensitivities to different conditions, we design a router aimed at adaptively evaluating and fusing these condition-specific representations. Specifically, we feed the patch features $f_I$ of the original image, the global image feature $f_I^g\in \mathbb{R}^{L\times d}$ obtained by expanding along the spatial dimension, and the concatenated multi-condition representation $C_{r}^I =\bigoplus_{i=1}^{N_c} R_{c_i}^I$ into the routing network to obtain the routing weights $\mathbf{W}^c\in \mathbb{R}^{L\times N_c}$ for each visual patch across all criteria, as shown below:

\begin{equation}
	\mathbf{W}^c = \sigma(\text{MLP}_{r}(f_I \oplus f_I^g \oplus C_{r}^I))
\end{equation}
Since the scene prototype and the original image correspond to the same underlying scene, the routing weights estimated from the original image are shared with the scene prototype to ensure criterion-consistent comparison. Accordingly, the weighted conditional representations are obtained as:
\begin{equation}
\hat{R}_{c_i}^x=R_{c_i}^x \odot \mathbf{W}^c_{:, i}, \qquad x\in\{I,I^*\}
\end{equation}

\subsubsection{Feature Injection}
The weighted condition features are concatenated and mapped through a GELU-activated MLP to produce a semantic increment, which is residually added to the patch feature. This process is formulated as:
\begin{equation}
	\hat{f}_x=f_x+\alpha  \cdot \text{MLP}_{f}(\bigoplus_{i=1}^{N_c} \hat{R}_{c_i}^x),\qquad x\in\{I,I^*\}
\end{equation}
where \(\alpha\) is a learnable residual scaling coefficient.

\subsection{Contrastive Decoding}
Finally, we fuse the conditionally augmented original patch feature $\hat{f}_{I}$, the conditionally augmented patch scene prototype feature $\hat{f}_{I^*}$, and the difference feature to form the visual conditional prefix $P_{v}$ for the LLM backbone. The difference feature highlights missing semantics. To avoid increasing sequence length through concatenation, we adopt residual fusion:

\begin{equation}
	P_{v} = \text{proj}(\hat{f}_{I}) + \gamma_p \cdot \text{proj}(\hat{f}_{I^*}) + \gamma_d \cdot (\text{proj}(\hat{f}_{I^*})-\text{proj}(\hat{f}_{I}))
\end{equation}
where $\gamma_p$ and $\gamma_d$ are trainable scalar variables initialized to 0, $\text{proj}(\cdot)$ is the pre-trained visual projection network designed to work with the LLM, and it is frozen throughout the entire training process. The LLM then generates negative captions describing the missing scene information in an autoregressive manner. The entire framework is trained end-to-end using four objectives: the anomaly classification loss $\mathcal{L}_a$, semantic alignment loss $\mathcal{L}_{m}$, existence loss $\mathcal{L}_{e}$, and caption generation loss $\mathcal{L}_{c}$, as shown below:
\begin{equation}
	\mathcal{L}_{total} = \mathcal{L}_a + \mathcal{L}_{m} + \mathcal{L}_{e} + \mathcal{L}_{c}
\end{equation}

\section{Experiment}

\begin{figure*}[!t]
	\centering
	\includegraphics[width=\linewidth]{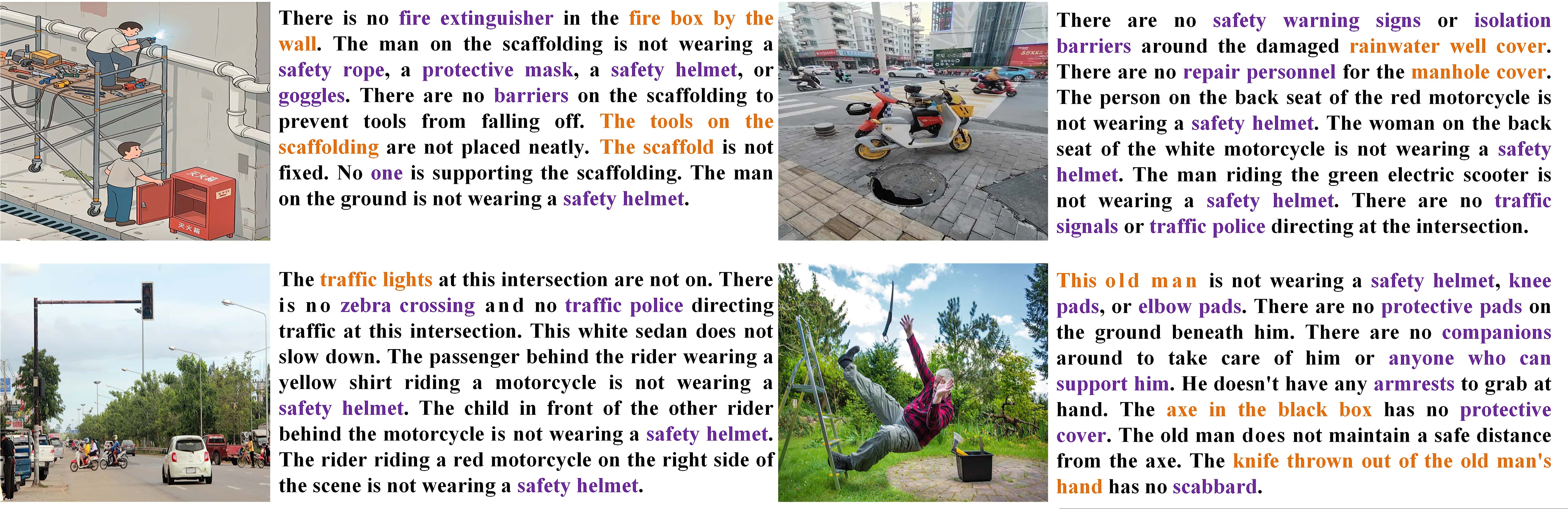}
	\caption{Some data examples from the SNUS dataset. Based on the original data, we add the object anomaly labels required for the Abnormal Judgment Module (abnormal objects are marked in \textcolor{orange}{orange}) and the missing object labels required for the Functional Association Branch (marked in \textcolor{purple}{purple}).}
	\label{data_example}
\end{figure*}

\subsection{Experiment Settings and Baselines}
\subsubsection{Dataset}
We evaluate our CRCD on the SNUS dataset, a benchmark introduced in our prior work specifically tailored for visual scene negative event understanding. This dataset comprises 12,118 instances, which are split into training, validation, and testing sets containing 8,770, 1,548, and 1,800 instances, respectively. It explicitly identifies key elements missing from images through purely negative text descriptions, and these absent elements directly correspond to potential safety risks in the scenes. The missing elements include objects, attributes, and relationships. However, the original SNUS dataset lacks the object anomaly labels for the Abnormal Judgment Module and the missing object labels for the Functional Association Branch. Therefore, we augment the dataset by constructing both sets of labels in this paper, as shown in Figure \ref{data_example}.

\subsubsection{Evaluation Metric}

We evaluate the quality of the generated negative captions using the CESG Score, which parses captions into counterfactual scene graphs and performs element-level matching. It reports precision, recall, and F1 scores for absent objects, attributes, and relationships, together with an overall CESG Score.

\subsubsection{Implementation Details}
To facilitate open-set object detection, CRCD incorporates Grounded SAM \cite{grounded_sam} to detect arbitrary objects without textual prompts. For experimental validation, we employ two representative MLLMs: LLaVA-1.5-7B \cite{llava} and Qwen2.5-VL-3B \cite{qwen25_vl}. Five safety-related criteria, including weather, light intensity, scene, crowd density, and protective equipment, are selected for multi-condition representation learning. For efficient parameter fine-tuning, we employ low-rank adaptation (LoRA)\cite{lora} for both the LLM in the functional association branch and the caption-decoding LLM, with the rank set to 8. The final CESG Score is computed via a weighted aggregation, with empirical coefficients assigned as 0.5, 0.25, and 0.25 for object, relation, and attribute components, respectively. All experiments are conducted using 2 NVIDIA A6000 GPUs. For more specific details, please refer to Appendix \ref{criteria_crl}.

\subsection{Method Comparison}
\subsubsection{Baseline Methods}
To comprehensively evaluate CRCD, we select representative baselines from five complementary perspectives: zero-shot capability, supervised task adaptation, preference alignment, explicit reasoning, and caption self-correction. These baselines examine whether visual negative captioning can be addressed by intrinsic MLLM knowledge, conventional supervised learning, text-level alignment, enhanced reasoning, or generic caption refinement. 1) Base model: directly uses off-the-shelf MLLMs without task-specific adaptation. 2) Vanilla Supervised Fine-Tuning (SFT)\cite{sft}: fine-tunes the base model using task-specific supervision. 3) Direct Preference Optimization (DPO)\cite{dpo}: evaluates whether preference alignment can mitigate affirmation bias. 4) Group Relative Policy Optimization (GRPO)\cite{grpo}: examines whether explicit textual reasoning can infer missing information without counterfactual visual reconstruction. 5) Odds Ratio Preference Optimization (ORPO)\cite{orpo}: provides a reference-free preference optimization baseline. 6) Cycle Consistency as Preference (CyclePref)\cite{cycle_preference}: constructs preference data using a cycle-consistency reward. 7) Self-Correction Caption (SC-Caption)\cite{sc_captioner}: evaluates whether generic self-correction can improve negative caption generation. Further implementation details are provided in Appendix~\ref{implement_detail}.

\subsubsection{Results Analysis}

Under the CESG evaluation framework, Table \ref{cesg_metric} presents a quantitative comparison between the proposed CRCD framework and the baseline methods. The unadapted base models exhibit particularly limited capability for this task. Qwen2.5-VL and LLaVA1.5 achieve only 27.04 and 10.25 CESG Scores, respectively, with especially low recall for missing attributes and relationships. This result indicates that general-purpose MLLMs rarely identify safety-critical non-existence spontaneously, despite their strong generic vision-language capabilities. After SFT, their CESG Scores increase substantially to 52.94 and 52.78, confirming that task-specific supervision can partially activate negative semantic generation. Nevertheless, the remaining gap to CRCD suggests that supervision alone is insufficient to overcome the intrinsic difficulty of reconstructing and grounding absent information. Overall, whether using the lightweight Qwen2.5-VL or the larger LLaVA1.5 as the base model, CRCD achieves the best performance on the comprehensive CESG Score. Specifically, CRCD reaches 68.89 on Qwen2.5-VL, outperforming the strongest competing baseline DPO (57.82), and achieves 68.73 on LLaVA1.5, exceeding CyclePref (55.11).

\begin{figure*}[!t]
	\centering
	\includegraphics[width=\linewidth]{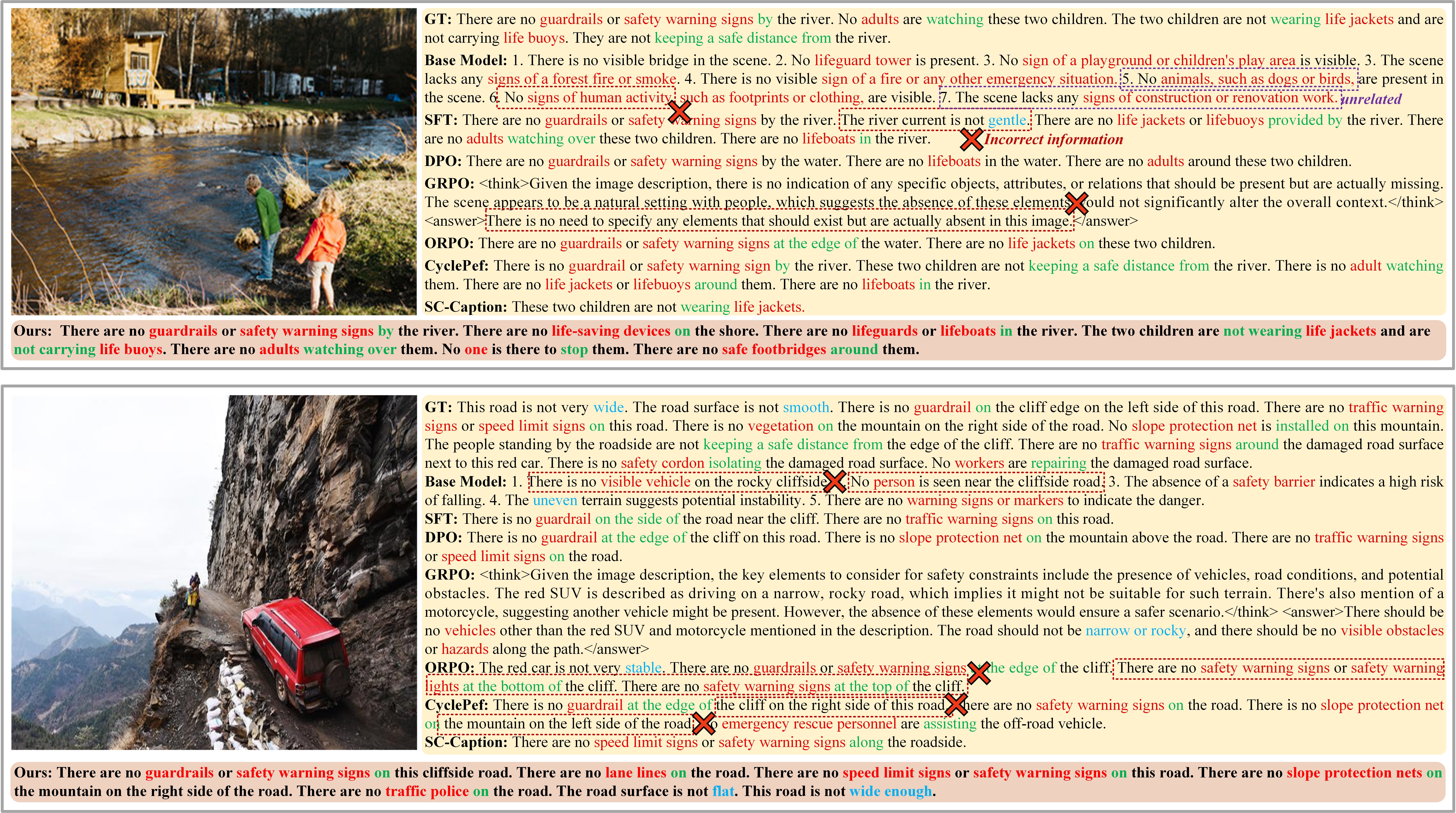}
	\caption{Examples of results under different methods. \textcolor{red}{Red}, \textcolor{blue}{blue}, and \textcolor{green}{green} are used to mark inferred missing objects, attributes, and relationships, respectively.}
	\label{result_example}
\end{figure*}

A detailed analysis of individual metrics reveals a distinct precision-recall trade-off. CRCD does not yield a significant precision advantage over existing baselines and even falls slightly behind SFT in some cases (e.g., \(P_o=93.69\) vs. 94.08 on Qwen2.5-VL). However, this marginal difference in precision is substantially compensated for by its clear advantage in recall, which primarily contributes to the improvement in the overall CESG Score. The poor recall performance of the baselines stems fundamentally from their limited mental filling capability. Without explicit modeling of missing information, they typically identify only a few prominent anomalies. Specifically, SC-Caption performs poorly on both base models, yielding CESG Scores of 23.61 on Qwen2.5-VL and 35.37 on LLaVA1.5. Driven by a self-correction mechanism that over-emphasizes conciseness, SC-Caption typically yields single-sentence outputs. This lack of detail sacrifices informational completeness, restricting its recall performance (e.g., \(R_o=16.59\) and 29.76 on the two models, respectively). On Qwen2.5-VL, GRPO also performs poorly with a CESG Score of 28.95, demonstrating that relying solely on text-space reasoning is insufficient for understanding visual negative events. Lacking the grounding of reconstructed visual features, GRPO's textual reasoning becomes poorly grounded in the visual scene, leading to logical collapse (e.g., the first example in Figure \ref{result_example}). Though CyclePref and DPO perform relatively well on Qwen2.5-VL (55.36 and 57.82, respectively) and LLaVA1.5 (55.11 and 55.07, respectively) via preference alignment, they remain limited to implicit text-level strategy optimization. Consequently, these methods fail to resolve the mental filling-in challenge at the level of underlying visual representations.

In contrast, CRCD achieves comprehensive performance gains by explicitly modeling safety-critical negation semantics at the feature level, translating directly into a decisive lead in the overall CESG Score. Specifically, on Qwen2.5-VL, CRCD improves the overall CESG Score to 68.89 (vs. 52.94 for SFT), backed by balanced F1-score improvements across $F1_o$, $F1_r$, and $F1_a$ to 75.32 (vs. 65.82), 61.48 (vs. 39.22), and 63.43 (vs. 40.91), respectively. Similarly, on LLaVA-1.5, CRCD elevates the CESG Score to 68.73 (vs. 52.78 for SFT), driven by sub-dimension F1-score leaps to 75.24 (vs. 66.58), 62.89 (vs. 39.65), and 61.55 (vs. 38.29). These results demonstrate that reconstructing visual expectations successfully mitigates both representation and affirmation biases, enabling the language decoder to capture fine-grained missing elements under strict safety criteria.

The qualitative results in Figure \ref{result_example} visually substantiate these findings. In the river scenario, baselines lacking visual anchors exhibit clear limitations: SC-Caption generates an overly brief description omitting critical safety elements, while GRPO experiences complete reasoning collapse, falsely concluding that no information is missing. In contrast, CRCD successfully retrieves the annotated guardrails and warning signs, while performing advanced common-sense inference to identify missing lifeguards and life-saving devices. In the mountain road scenario, while baselines like DPO and CyclePref miss fine-grained safety attributes (e.g., lane lines and speed limit signs), and ORPO generates physically contradictory warning sign locations, CRCD precisely reconstructs the spatial-semantic scene prototype to recover all key missing hazards. This demonstrates CRCD's robust capability to bridge physical reality with logical expectations.

\begin{table*}
	\caption{Performance Comparison of Different Methods Under CESG Evaluation Framework.}
	\centering
		\begin{tabular}{ccccccccccc}
			\hline
			\toprule
			
			\textbf{Method}&$\mathbf{P_o}$&$\mathbf{R_o}$&$\mathbf{F1_o}$&$\mathbf{P_r}$&$\mathbf{R_r}$&$\mathbf{F1_r}$&$\mathbf{P_a}$&$\mathbf{R_a}$&$\mathbf{F1_a}$ &\textbf{CESG Score} \\
			\midrule
			\multicolumn{11}{c}{\cellcolor{gray!15}\textbf{Base Model: Qwen2.5-VL-3B}} \\
			\midrule
			Base Model\cite{qwen25_vl}   & 66.32 & 26.07 & 37.43 & 66.24 & 20.08 & 30.82 & 69.86 & 1.27 & 2.49  & 27.04 \\
			SFT\cite{sft}    & \textbf{94.08} & 50.62 & 65.82 & \textbf{84.06} & 25.57 & 39.22 & \underline{79.21} & 27.57 & 40.91  & 52.94 \\
			DPO\cite{dpo}    & 87.43 & \underline{54.12} & \underline{66.86} & 50.59 & \underline{48.78} & 49.67 & 76.37 & \underline{34.89} & \underline{47.90} & \underline{57.82} \\
			GRPO \cite{grpo}   & 54.72 & 24.13 & 33.49 & 38.11 &  17.30 &23.80 & 43.49 &  17.54 & 25.00  & 28.95 \\
			ORPO \cite{orpo}   & 91.15 & 31.95 & 47.31 & 81.62 & 16.28 & 27.15 & 77.23 & 25.21 & 38.01  & 39.95 \\
			SC-Caption \cite{sc_captioner}   & 92.39 & 16.59 & 28.13 & 79.58 & 10.55 & 18.63 & 71.34 & 11.32 & 19.54  & 23.61 \\
			CyclePref \cite{cycle_preference}   & 87.66 & 51.92 & 65.21 & 54.22 & 46.33 & \underline{49.96} & 68.35 & 29.32 & 41.04  & 55.36 \\
			CRCD (Ours)    & \underline{93.69} &\textbf{62.97} & \textbf{75.32} & \underline{82.52} &\textbf{48.99} &\textbf{61.48} & \textbf{80.97} & \textbf{52.13}& \textbf{63.43} & \textbf{68.89}\\
			\midrule
			\multicolumn{11}{c}{\cellcolor{gray!15}\textbf{Base Model: LLaVA1.5-7B}} \\
			\midrule
			Base Model\cite{llava} & 65.40& 5.95  & 10.90 & 62.65 & 5.15 & 9.52 & 49.38 & 3.09 & 5.82 & 10.25 \\
			SFT\cite{sft}  & \textbf{95.40}& 51.13  & \underline{66.58} & \textbf{84.10} & 25.94 & 39.65 & 81.77 & 25.00 & 38.29 & 52.78 \\
			DPO\cite{dpo}  & 80.10& 55.40  &65.50 &69.74& 35.54 & 47.09 & 52.56&35.25 & 42.20 & 55.07 \\
			ORPO  \cite{orpo}  & 92.29 & 36.91 & 52.73 & 81.20 & 19.00 & 30.80 & \textbf{83.37} & \underline{35.72} & \underline{50.01}  &  46.57\\
			SC-Caption \cite{sc_captioner}   & 79.50 & 29.76 & 43.31 & 74.14 & 18.76 & 29.94 & 64.80 & 15.42 & 24.91  & 35.37 \\
			CyclePref \cite{cycle_preference}   & 78.88 & \underline{56.71} & 65.98 & 55.41 & \underline{44.65} & \underline{49.45} & 67.41 & 27.47 & 39.03  & \underline{55.11} \\
			CRCD (Ours)    & \underline{95.02} & \textbf{62.27} & \textbf{75.24} & \underline{83.54} & \textbf{50.42} & \textbf{62.89} &\underline{82.38}& \textbf{49.13}& \textbf{61.55} & \textbf{68.73} \\
			\bottomrule
		\end{tabular}
	\label{cesg_metric}
\end{table*}

\subsection{Ablation Study}

\subsubsection{Ablation Study on the Core Modules}
To validate the contribution of each core component in CRCD, we conduct systematic ablation studies. By progressively integrating the Amodal Completion Branch (ACB), Functional Association Branch (FAB), and Multi-Condition Representation Learning (MCRL), we quantitatively analyze their individual impacts on the understanding of missing objects, attributes, and relationships. The results are summarized in Table \ref{ablation_module}.

\begin{table*}[t]
	\centering
	\caption{Module ablation study for our proposed framework. \textbf{ACB}, \textbf{FAB}, and \textbf{MCRL} denote the Amodal Completion Branch, Functional Association Branch, and Multi-Condition Representation Learning, respectively.}
	\label{ablation_module}
		\begin{tabular}{ccc| c|ccc ccc ccc c}
			\toprule
			\textbf{ACB} & \textbf{FAB} & \textbf{MCRL}  &  \textbf{\#Trainable Params}&$\mathbf{P_o}$& $\mathbf{R_o}$ & $\mathbf{F1_o}$ & $\mathbf{P_r}$ & $\mathbf{R_r}$ & $\mathbf{F1_r}$ & $\mathbf{P_a}$ & $\mathbf{R_a}$ & $\mathbf{F1_a}$ & \textbf{CESG Score}\\
			\midrule
			\multicolumn{14}{c}{\cellcolor{gray!14}\textbf{Base Model: Qwen2.5-VL-3B}} \\
			\midrule
			&            &            & 4.72M &94.08 & 50.62 & 65.82 & 84.06 & 25.57 & 39.22 & 79.21 & 27.57 & 40.91& 52.94 \\
			\checkmark	&   &            & 46.02M &91.73& 60.76& 73.10 & \underline{85.01} & 47.26 & \underline{60.75} & 78.63 & 42.13 & 54.86 & 65.45\\
			&  \checkmark    &            &49.75M &\textbf{95.01} &61.24 & 74.48 &82.08 &\underline{47.95} &60.54 & 79.93 & 46.63& 58.90 & 67.10\\
			\checkmark & \checkmark &            & 52.52M &92.94 & \underline{62.86} & \underline{75.00} & \textbf{85.09} & 43.56 & 57.62 &\textbf{81.10} & \underline{49.24} & \underline{61.28} & \underline{67.23} \\
			&            & \checkmark &6.67M &\underline{94.63} & 52.37 & 67.43 & 82.93 & 40.55 & 54.47 & 80.34 & 42.79 & 55.84 & 61.29\\
			\checkmark & \checkmark & \checkmark &54.47M &93.69 &\textbf{62.97} & \textbf{75.32} & 82.52 &\textbf{48.99} &\textbf{61.48} & \underline{80.97} & \textbf{52.13}& \textbf{63.43} & \textbf{68.89}\\
			\midrule
			\multicolumn{14}{c}{\cellcolor{gray!14}\textbf{Base Model: LLaVA1.5-7B}} \\
			\midrule
			&            &            &8.39M &95.40 &51.13 & 66.58 & \textbf{84.10} &25.94 &39.65 & 81.77 & 25.00& 38.29 & 52.78 \\
			\checkmark &  &           & 102.26M&\textbf{96.64} & 58.09 &72.56 & 82.37 & 44.29 & 57.61 & 78.74 &44.89 & 57.18 & 64.98 \\
			&     \checkmark       &          	& 115.10M  &95.88& \underline{60.31} & \underline{74.04} & 81.47 & 47.49 & 60.00 & 79.05 & 39.11 &52.33 & 65.10\\
			\checkmark & \checkmark &            & 118.15M &94.69 & 59.67 & 73.21 & 82.20 &  \underline{47.99} & \underline{60.60} &80.79& \underline{45.04} & \underline{57.84} & \underline{66.22} \\
			&            & \checkmark & 10.06M &\underline{96.32} & 52.47 & 67.93 & 82.74 & 45.86 & 59.01 & \textbf{83.25} & 43.25& 56.93 & 62.95 \\
			\checkmark & \checkmark & \checkmark & 119.82M &95.02 & \textbf{62.27} & \textbf{75.24} & \underline{83.54} & \textbf{50.42} & \textbf{62.89} & \underline{82.38}& \textbf{49.13}& \textbf{61.55} & \textbf{68.73} \\
			\bottomrule
		\end{tabular}%
	\vspace{-0.15in}
\end{table*}

The experimental results demonstrate that the dual-branch architecture plays a critical role in handling negative events at distinct cognitive levels. Introducing either ACB or FAB alone significantly improves upon the no-module configuration (where training under this setup is equivalent to standard SFT on the base model), primarily driven by substantial recall gains. This indicates that ACB and FAB successfully restore incomplete object features and infer non-existent objects, respectively, establishing robust visual anchors for the language decoder. Notably, FAB yields a greater CESG Score improvement than ACB (67.10 vs. 65.45 on Qwen2.5-VL and 65.10 vs. 64.98 on LLaVA1.5, compared to the no-module baseline scores of 52.94 and 52.78, respectively). This gap stems from their distinct cognitive dimensions. ACB implicitly recovers local features of incomplete but detected objects. Since pre-trained VLMs already possess inherent fault tolerance for such incomplete semantics via their robust vision-language alignment, the additional gains from ACB are naturally bounded. In contrast, FAB targets counterfactual non-existent objects that leave no pixel-level cues, directly addressing the model's limited capability to infer missing entities through functional association. By reconstructing these missing objects in the feature space, FAB bypasses the perceptual blind spots of the base model, unlocking greater performance potential. Integrating both branches yields synergistic gains (67.23 on Qwen2.5-VL and 66.22 on LLaVA1.5) by enriching the semantic completeness of the scene prototypes.

The MCRL module also brings independent improvements over the no-module baseline (raising the CESG Score to 61.29 on Qwen2.5-VL and 62.95 on LLaVA1.5) by mitigating representation bias. By decoupling and adaptively recoupling visual features under safety-critical criteria, MCRL forces the model to dynamically prioritize multi-dimensional safety information across different local regions. This mechanism implicitly acts as a safety-oriented visual question answering prior \cite{crl} before autoregressive decoding. When MCRL is integrated with the dual-branch framework (ACB + FAB), the full configuration achieves the optimal performance (68.89 on Qwen2.5-VL and 68.73 on LLaVA1.5). This peak performance indicates that while the dual branches construct structurally sound scene prototypes, MCRL injects essential sensitivity to safety-related guidelines, enhancing the fine-grained semantic detail of the generated captions.

While CRCD achieves decisive performance gains, it entails increased computational complexity. As shown in Table \ref{ablation_module}, trainable parameters scale from the no-module baseline (4.72M for Qwen2.5-VL and 8.39M for LLaVA1.5) to the full configuration (54.47M and 119.82M, respectively). This parameter growth constitutes a current limitation: reconstructing high-fidelity safety scene prototypes in the latent space requires additional feature generation, spatial projection, and fusion modules, which inevitably increase training-phase GPU memory overhead.

\subsubsection{Ablation Study on the Loss Functions}
To verify the contributions of each loss function, we conduct an incremental ablation study on the joint objective function. Specifically, we use the initial configuration relying solely on the language generation loss $\mathcal{L}_c$ as our baseline, and progressively introduce the other three loss terms. The results are shown in Table \ref{ablation_loss}.

\begin{table*}[t]
	\centering
	\caption{Loss function ablation study for our proposed framework.}
	\label{ablation_loss}
		\begin{tabular}{cccc|ccc ccc ccc c}
			\toprule
			$\mathcal{L}_c$ & $\mathcal{L}_a$ & $\mathcal{L}_m$ & $\mathcal{L}_e$ &$\mathbf{P_o}$& $\mathbf{R_o}$ & $\mathbf{F1_o}$ & $\mathbf{P_r}$ & $\mathbf{R_r}$ & $\mathbf{F1_r}$ & $\mathbf{P_a}$ & $\mathbf{R_a}$ & $\mathbf{F1_a}$ & \textbf{CESG Score}\\
			\midrule
			\multicolumn{14}{c}{\cellcolor{gray!14}\textbf{Base Model: Qwen2.5-VL-3B}} \\
			\midrule
			\checkmark	&   &            && 91.86 & 57.93& 71.05 & 78.83 &46.21& 58.27& 78.84 & 47.92 & 59.61& 65.00 \\
			\checkmark&  \checkmark    &      &      & 92.33& 61.87& 74.09 & 79.05 & 46.87 & 58.85 & 78.32& 48.16 & 59.64 & 66.67 \\
			\checkmark & \checkmark & \checkmark && \underline{92.76} & \bf 63.09 &\underline{75.10} & \underline{80.18} & \underline{48.86} & \underline{60.72} & \underline{80.06} & \underline{51.77}& \underline{62.88} & \underline{68.45}\\
			\checkmark & \checkmark & \checkmark & \checkmark& \bf 93.69 & \underline{62.97} & \bf 75.32 & \bf 82.52 & \bf 48.99 &\bf 61.48 & \bf 80.97 & \bf 52.13& \bf 63.43 & \bf 68.89\\
			\midrule
			\multicolumn{14}{c}{\cellcolor{gray!14}\textbf{Base Model: LLaVA1.5-7B}} \\
			\midrule  
			\checkmark	&   &            &&\underline{93.89}& 58.99& 72.46 & 81.38 & 48.02& 60.40 & 79.29 & 44.50 & 57.01 & 65.58 \\
			\checkmark&  \checkmark    &      &      & 93.43 &60.29& 73.63 & 81.07 & 48.04 & 60.33& \underline{81.50} & 44.89 & 57.89& 66.37 \\
			\checkmark & \checkmark & \checkmark && 93.59 & \bf 63.68& \bf 75.79 & \underline{81.69} & 
			\underline{49.07} &\underline{61.31} & 80.97 & \bf 50.05& \bf 61.86 & \underline{68.69}\\
			\checkmark & \checkmark & \checkmark & \checkmark& \bf 95.02 & \underline{62.27} & \underline{75.24} & \bf 83.54 & \bf 50.42 & \textbf{62.89} & \textbf{82.38}& \underline{49.13}& \underline{61.55} & \textbf{68.73} \\
			\bottomrule
		\end{tabular}%
	\vspace{-0.15in}
\end{table*}

Under the initial setup, Qwen2.5-VL and LLaVA1.5 achieve overall CESG scores of 65.00 and 65.58, respectively. Incorporating the classification loss $\mathcal{L}_a$ for anomaly judgment significantly boosts performance, raising the CESG scores to 66.67 and 66.37. This improvement is primarily driven by a sharp increase in object recall ($R_o$). Explicitly penalizing misclassifications of object states enables the gated visual features to more accurately pinpoint anomalous regions, providing high-quality spatial anchors for subsequent amodal completion. Building on this, introducing the semantic alignment loss $\mathcal{L}_m$ for the functional association branch leads to a balanced and robust improvement across all recall metrics, further elevating the CESG scores to 68.45 and 68.69. This loss guides the learnable query tokens to reconstruct diverse and unique counterfactual features, extending the semantic coverage of the reconstructed safety risks. Finally, the existence supervision term $\mathcal{L}_e$ completes our joint objective framework, yielding the optimal overall performance (68.89 on Qwen2.5-VL and 68.73 on LLaVA1.5). By dynamically suppressing the confidence scores of redundant or mismatched query tokens, $\mathcal{L}_e$ effectively reduces spurious counterfactual predictions and prevents the decoder from describing unsupported missing objects.

\subsubsection{Ablation Study on the Number of Queries}
In the Functional Association Branch, the number of learnable query tokens $N_q$ determines the capacity of the model to associate completely missing objects. To investigate the impact of $N_q$ on the generated negative captions, we conduct ablation experiments on both base models by setting $N_q$ to 1, 5, 10, and 20. The results are summarized in Table \ref{ablation_crl_query_num}.

\begin{table*}[t]
	\centering
	\caption{Ablation study on the number of queries in the functional association branch.}
	\label{ablation_crl_query_num}
		\begin{tabular}{c |ccc ccc ccc c}
			\toprule
			\textbf{The Number of Queries}& $\mathbf{P_o}$ & $\mathbf{R_o}$ & $\mathbf{F1_o}$ & $\mathbf{P_r}$ & $\mathbf{R_r}$ & $\mathbf{F1_r}$ & $\mathbf{P_a}$ & $\mathbf{R_a}$ & $\mathbf{F1_a}$ & \textbf{CESG Score}\\
			\midrule
			\multicolumn{11}{c}{\cellcolor{gray!11}\textbf{Base Model: Qwen2.5-VL-3B}} \\
			\midrule
			$N_q=1$ & \textbf{94.50}& 54.82& 69.39 & \textbf{83.58} & 40.12& 54.22& \underline{80.38} & 46.55 & 58.96& 62.99 \\
			$N_q=5$ &\underline{93.69} &62.97 & \textbf{75.32} & \underline{82.52} &48.99 &\textbf{61.48} & \textbf{80.97} & \textbf{52.13}& \textbf{63.43} & \textbf{68.89} \\
			$N_q=10$  &91.95 & \underline{63.25}& \underline{74.95} & 81.94 & \underline{49.14}& \underline{61.44}& 80.16 & 51.94 &\underline{63.04}& \underline{68.60} \\
			$N_q=20$ &89.83 & \textbf{63.28} & 74.25 & 80.57 & \textbf{49.39} & 61.24 & 78.94 & \underline{52.05} & 62.73 & 68.12\\
			
			\midrule
			\multicolumn{11}{c}{\cellcolor{gray!11}\textbf{Base Model: LLaVA1.5-7B}} \\
			\midrule
			$N_q=1$ &\underline{94.75} & 54.90 & 69.52 & 82.64 & 45.18 & 58.42 & \textbf{82.47} & 43.20 & 56.70& 63.54 \\
			$N_q=5$ &\textbf{95.02} & 62.27 & \textbf{75.24} & \underline{83.54} & \underline{50.42} & 62.89 & \underline{82.38}& 49.13& 61.55 & 68.73 \\
			$N_q=10$ &94.33 & \underline{62.43} & \underline{75.13} & \textbf{83.98} &50.36 & \underline{62.96} &82.09 & \underline{50.28} & \underline{62.36} & \underline{68.90} \\
			$N_q=20$ &92.19 & \textbf{63.07} & 74.90 & 81.21 & \textbf{51.76} & \textbf{63.22} & 81.97 & \textbf{50.73} & \textbf{62.67} & \textbf{68.92}\\
			\bottomrule
		\end{tabular}%
	\vspace{-0.15in}
\end{table*}

As $N_q$ increases from 1 to 5, the model exhibits a substantial overall improvement, primarily driven by recall gains. When $N_q=1$, the model's capacity is highly restricted, only capturing the single most prominent missing object and yielding a CESG Score of 62.99 on Qwen2.5-VL and 63.54 on LLaVA1.5. Expanding $N_q$ to 5 substantially boosts recall metrics, driving the CESG Scores to 68.89 and 68.73, respectively. This performance gain aligns with the statistical distribution of the SNUS dataset, where multiple missing elements typically coexist in safety-critical scenarios. Increasing the number of query vectors effectively resolves this representation bottleneck, enabling the framework to concurrently retrieve and reconstruct multiple counterfactual objects in the latent space.

However, further increasing $N_q$ to 10 and 20 leads to a clear performance plateau. While recall metrics show marginal improvements, the overall precision tends to decline. For instance, on Qwen2.5-VL, the object precision $P_o$ declines from 93.69 to 91.95 and 89.83, while on LLaVA1.5, it drops from 95.02 to 94.33 and 92.19. This trade-off stems from an oversized query space introducing redundant representation slots. Although the framework incorporates an existence head for soft filtering, the surplus of latent queries increases optimization noise during Hungarian matching, which ultimately misleads the language decoder into generating hallucinated descriptions of irrelevant objects outside the strict safety constraint.

\subsection{Probe-based Verification of Visual Reconstruction}

To verify whether the reconstructed latent scene prototype and the generated counterfactual visual vectors effectively capture the semantic features of missing objects, we conduct three probe-based verification experiments. These experiments serve to decouple the internal representation from the final language generation, providing evidence that the performance gains arise from meaningful visual reconstruction rather than solely from linguistic generation.

\subsubsection{Feature Completion Visualization via t-SNE\cite{tsne}}

\begin{figure}[!t]
	\centering
	\includegraphics[width=\linewidth]{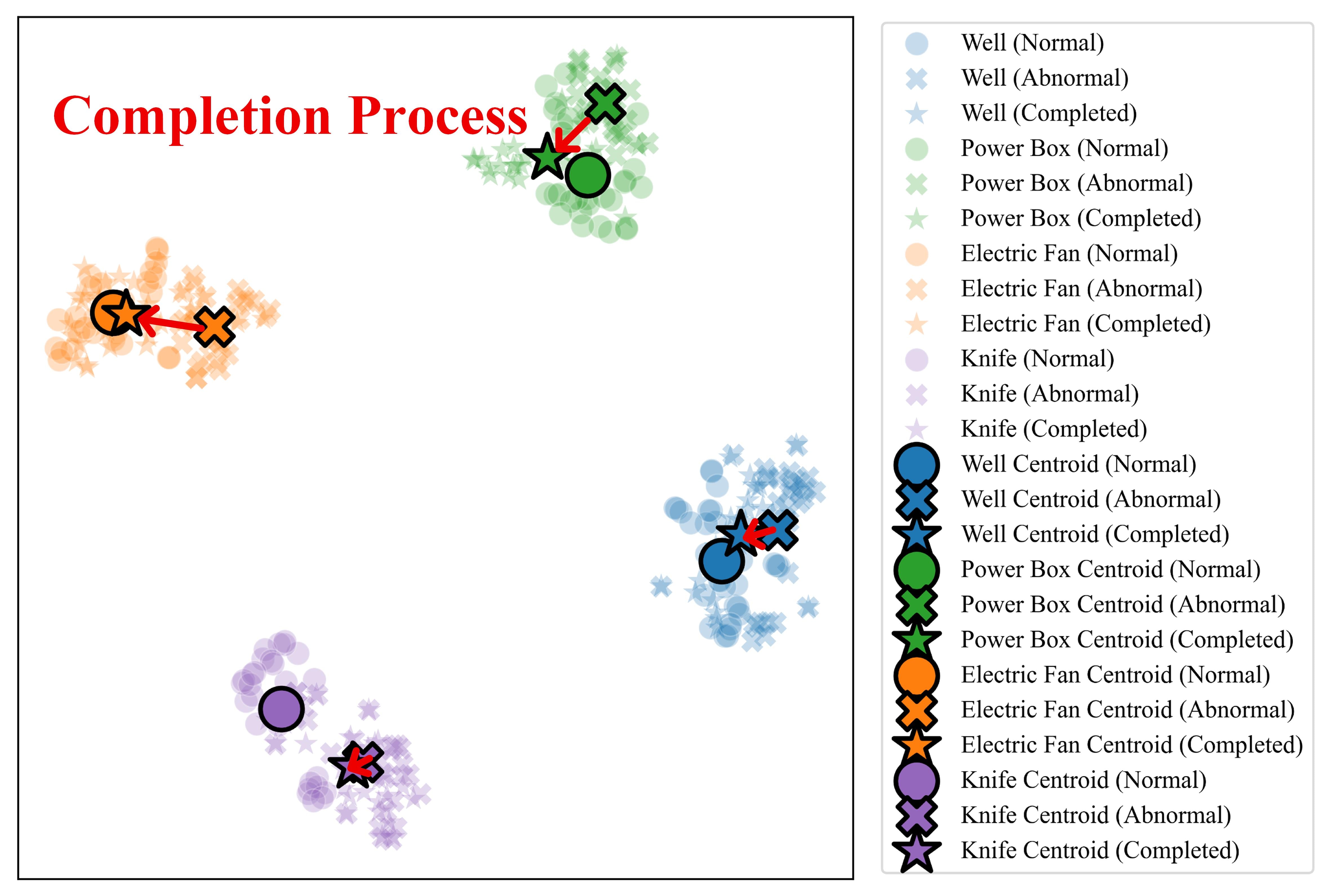}
	\caption{t-SNE visualization of feature manifolds and centroid trajectory alignment via the Amodal Completion Branch.}
	\label{first_branch_results}
\end{figure}

We first applied the t-SNE algorithm to visualize the features in the amodal completion branch. We select four common object categories whose missing parts may pose safety risks. We visualize their sample embeddings and cluster centroids in a shared low-dimensional space, covering normal, abnormal, and completed states, as shown in Figure \ref{first_branch_results}.
 As shown in Figure \ref{first_branch_results}, we select four common object categories whose missing parts may pose safety risks. We visualize their sample embeddings and cluster centroids in a shared low-dimensional space, covering the normal states, abnormal states, and the states of abnormal samples after completion by the Amodal Completion Branch. 
 These four object categories are \textit{well}, \textit{power box}, \textit{knife}, and \textit{electric fans}, with corresponding abnormal states being missing manhole covers, missing box doors, missing knife sheaths, and missing protective covers. The results show that the cluster centroids of the anomalous samples undergo a shift after passing through the amodal completion branch, reducing the distance between them and the cluster centroid of samples in the normal state. This indicates that the branch is capable of extracting missing semantic information from text embeddings and effectively integrating it into the visual representations of the anomalous objects.

\subsubsection{Feature Alignment Probe via CLIP Similarity}

\begin{figure*}[h]
	\centering
	\subfloat[]{\includegraphics[width=0.915\linewidth]{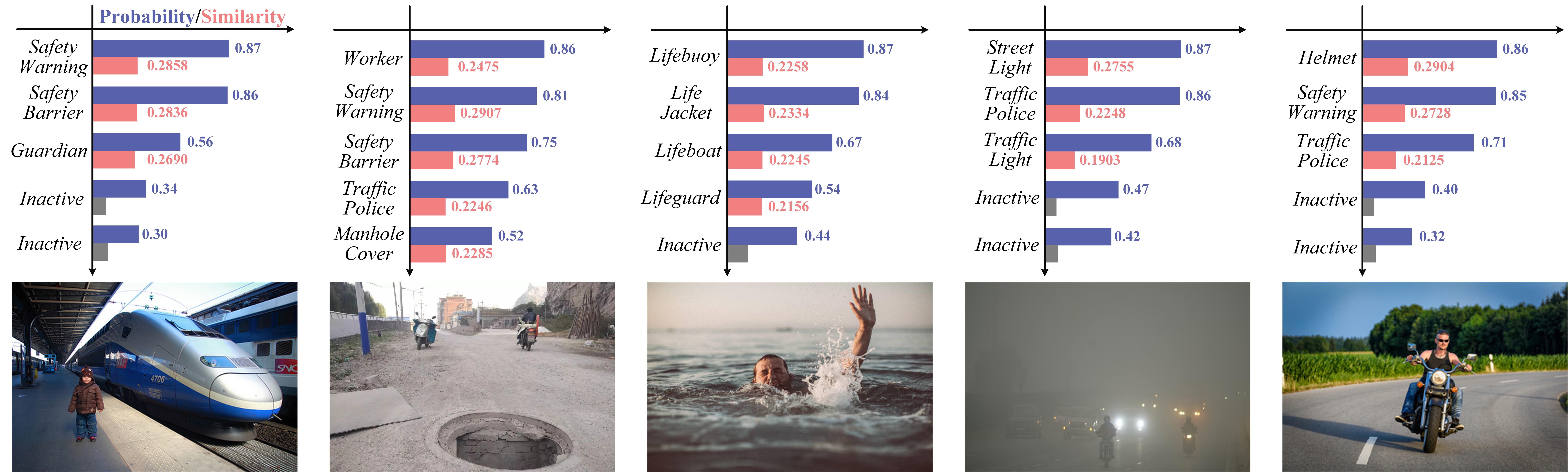}%
		\label{ori_images_and_clip_sim}}
	\hfil
	\subfloat[]{\includegraphics[width=0.9\linewidth]{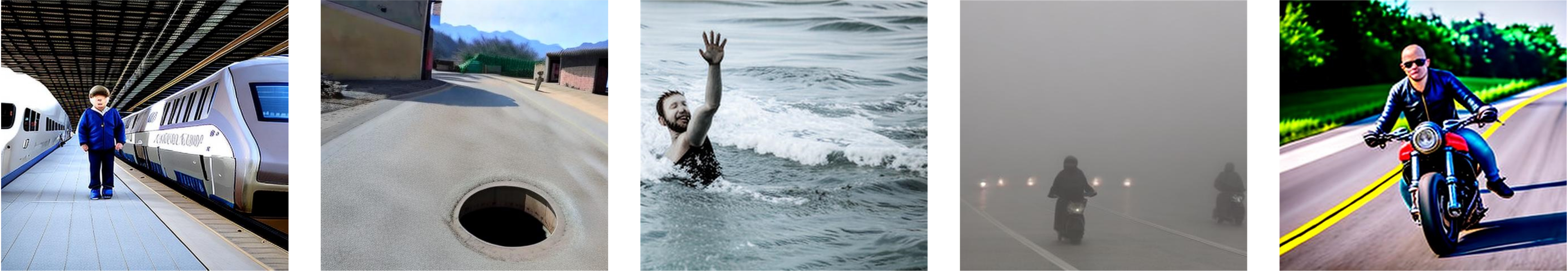}%
		\label{ori_generated_images}}
	\hfil
	\subfloat[]{\includegraphics[width=0.9\linewidth]{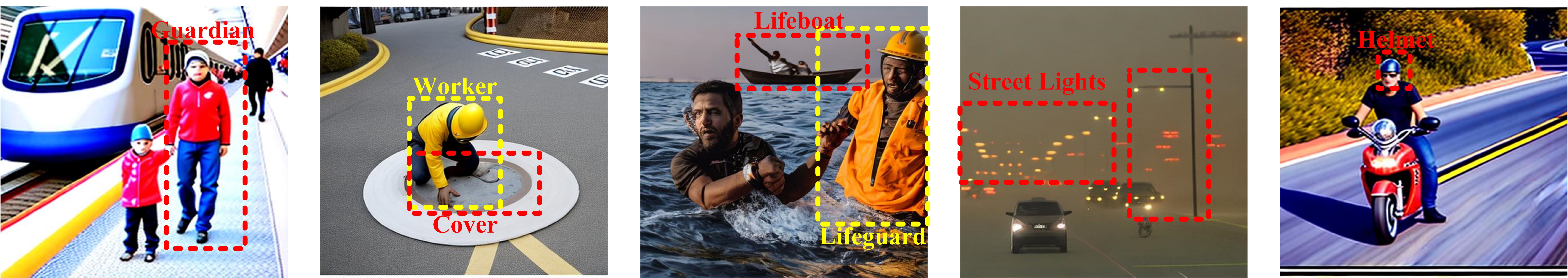}%
		\label{scene_prototye_generated_images}}
	\caption{Probe Verification Results. (a) Original image and the retrieval status of the associated objects in the offline library. (b) Generated image based on the original image feature. (c) Generated image based on the scene prototype feature.}
	\label{scene_prototype}
\end{figure*}

We quantitatively examine whether the counterfactual vectors $\{v_i^*\}_{i=1}^{N_q}$ generated by the functional association branch encode meaningful missing-object semantics. Specifically, we project the active vectors into the CLIP-aligned embedding space and retrieve their nearest text entries from a predefined safety-object library. As shown in Figure \ref{ori_images_and_clip_sim}, the retrieved entries are semantically consistent with the intended missing objects, with cosine similarities mostly ranging from 0.2 to 0.3. Although these moderate similarities do not imply high-fidelity reconstruction, they indicate that the inferred vectors encode meaningful object-level semantics rather than arbitrary noise. Queries with expected-existence confidence below 0.5 are suppressed, allowing the fixed query set to produce a variable number of active missing-object predictions.

\subsubsection{Visual Semantic Probe via unCLIP}

This experiment aims to reconstruct the scene prototype into pixel space to intuitively assess whether the inferred missing information is present. As a technique that relies solely on image embeddings to generate corresponding images, unCLIP\cite{unclip} is compatible with the CLIP-based visual representation used by LLaVA1.5. Therefore, we introduce the unCLIP technique as a visual probe, feeding the scene prototype feature directly into the pre-trained unCLIP diffusion decoder to obtain a generated image based on the scene prototype feature, as illustrated in Figure\ref{scene_prototye_generated_images}. Concurrently, we also input the feature of the original image into the pre-trained unCLIP model to obtain a generated image based on the original feature, serving as the baseline, as illustrated in Figure\ref{ori_generated_images}. Although unCLIP cannot achieve high-fidelity reconstruction of images, we can still intuitively observe that the generated images based on scene prototype features contain missing information elements that are absent in the original images. This indicates that CRCD has successfully injected the feature information of the missing object into the scene prototype in a substantive manner.

\section{Conclusion}
In this paper, we introduce CRCD, a framework that reformulates visual negation understanding into a latent change-captioning process. Grounded in the psycholinguistic two-step simulation hypothesis, CRCD reconstructs expected but absent safety elements as a counterfactual latent scene prototype, subsequently generating negative captions via contrastive decoding. Extensive evaluations on the SNUS dataset demonstrate that CRCD effectively mitigates representation bias and suppresses hallucinations when reasoning about absent objects, attributes, and relationships. Overall, this work highlights the value of top-down contrastive reasoning for safety-oriented scene understanding and provides a strong baseline for visual negative captioning.

\bibliographystyle{IEEEtran}
\bibliography{reference}

\clearpage 
\setcounter{page}{1}
{
\appendices

\section{Additional Implementation Details}
\subsection{Detailed Implementation Instructions for the Comparison Methods}
\label{implement_detail}

The prompt template used in the base models and SFT method is as follows: "\textit{Constrained by safety cognition, this task relies on information visible at the pixel level within a scene to generate coherent textual paragraphs describing elements that should exist but are actually absent. This includes non-existent objects, non-existent attributes, and non-existent relationships. These absent elements characterize certain hazards or safety risks within the scene.  Their presence would improve the safety of the scene. Each sentence within the paragraph must be formulated as a negative statement.}"

DPO is implemented within the Llama Factory framework(\url{https://github.com/hiyouga/LlamaFactory}). We first train a base model using supervised data, and then construct preference data pairs to train the fine-tuned model. In the preference data, the chosen response is purely negative text, while the rejected response is purely affirmative text with the negative words removed. Similarly, we use the LoRA method for training, with LoRA configurations consistent with those of the SFT method. The implementation of the ORPO method is similar to that of DPO, with the difference being that the ORPO method skips the SFT step and proceeds directly to preference training.

We utilize the EasyR1 framework(\url{https://github.com/hiyouga/EasyR1}) to implement the GRPO algorithm. Since the LLaVA1.5 model lacks native reasoning capabilities, this method is implemented only on the Qwen model. We design a prompt to encourage explicit reasoning before generating negative captions, as follows: \textit{You FIRST think about the reasoning process as an internal monologue and then provide the final negative caption. The reasoning process MUST BE enclosed within $< \text{think} >$  $< \text{/think} >$ tags. The final negative caption MUST BE enclosed within $< \text{answer} >$  $< \text{/answer} >$ tags.} We also fine-tuned the model using the LoRA approach, with the rank set to 16. Additionally, the \textit{num\_generations} parameter for this method is set to 5, and the sampling temperature is set to 0.6.

As for CyclePref, we first construct the preference dataset based on the official source code for CycleReward(\url{https://github.com/hjbahng/cyclereward}). We then use this dataset to perform DPO training on the SFT model within the Llama Factory framework. The Llama Factory version included in the official SC-Caption source code(\url{https://github.com/zl2048/SC-Captioner}) does not support the Qwen 2.5 series of models. We have modified the latest version of Llama Factory to make it compatible with Qwen 2.5-VL. All other experimental parameters use the default settings.

\subsection{The Implementation Details of Our Method}
\label{criteria_crl}
\renewcommand{\tabularxcolumn}[1]{m{#1}}

For both LoRA adapters, we set the target modules to $q_{proj}$, $v_{proj}$, $k_{proj}$ and $o_{proj}$. We reuse the visual encoder and LLM backbone of each MLLM within the CRCD framework. For the LLaVA1.5-7B, the visual encoder uses the pre-trained CLIP-VIT-L/14-336px, so it has an off-the-shelf text encoder that is aligned with the visual encoder. In contrast, Qwen2.5-VL-3B employs a model-specific vision encoder and does not provide an off-the-shelf text encoder aligned with the same visual embedding space. To address this issue, we construct an embedding model based on the GME approach\cite{gme} before entering the training pipeline of the method framework. This embedding model then serves as the text encoder within the method framework. Additionally, in multi-condition representation learning, each criterion contains different values, as detailed in Table \ref{criterion_value}.

We adopt a two-phase training strategy. In the first phase, we freeze the two LoRA adapters and train the modules requiring training with a learning rate of $10^{-4}$. In the second phase, we unfreeze these two LoRA adapters and train them with a learning rate of $10^{-4}$, while reducing the learning rate for the remaining modules to $10^{-5}$.

\begin{table*}
	\caption{The criteria and corresponding values involved in multi-condition representation learning}
	\centering
	\begin{tabularx}{\textwidth}{ >{\centering\arraybackslash}m{3cm} >{\raggedright\arraybackslash}X }
		\toprule
		
		\textbf{Criterion} & \multicolumn{1}{c}{\textbf{Value}} \\ 
		\midrule
		
		Weather
		& 'rainy', 'sunny', 'clear', 'cloudy', 'overcast', 'snowy', 'windy', 'foggy', 'frosty', 'hazy', 'dry', 'warm', 'cold', 'hot', 'cool', 'humid' \\
		\midrule
		
		Light Intensity
		& 'brightness', 'bright illumination', 'harsh light', 'glaring light', 'strong light intensity', 'brilliant ambient light', 'full daylight', 'blazing sunlight', 'soft diffused light', 'moderate light intensity', 'even lighting', 'gentle illumination', 'natural daylight', 'mild light intensity', 'soft overhead lighting', 'dim lighting', 'gloomy light', 'shadowy lighting', 'poorly lit scene', 'faint illumination', 'subdued lighting', 'overexposed lighting', 'extremely low light', 'darkness', 'near-total darkness' \\
		\midrule
		
		Protective Equipment
		& 'safety helmet', 'hard hat', 'protective cap', 'safety goggles', 'protective glasses', 'face shield', 'safety visor', 'face mask', 'medical mask', 'surgical mask', 'N95 respirator', 'dust mask', 'gas mask', 'protective gloves', 'work gloves', 'rubber gloves', 'disposable gloves', 'latex gloves', 'protective clothing', 'coveralls', 'lab coat', 'safety vest', 'work uniform', 'earplugs', 'earmuffs', 'safety shoes', 'steel-toe boots', 'protective boots' \\
		\midrule
		
		Scene
		& 'outdoor scene', 'indoor scene', 'urban street', 'public place', 'hospital', 'open area', 'busy downtown', 'indoor room', 'mountainous area', 'quiet park', 'crossroads', 'expressway', 'construction site', 'roof', 'military battlefield', 'indoor hall', 'kitchen', 'living room', 'stairwell', 'savanna', 'wilderness', 'factory', 'workshop', 'school', 'bathroom', 'meeting room', 'office', 'overpass', 'a busy city', 'waterside', 'accident area', 'sports field', 'construction zone', "no man's land", "mountainous region", 'airport', 'a high altitude platform', 'edge zone', 'conflict zone', 'volcano', 'sidewalk' \\
		
		\midrule
		Crowd Density
		& 'sparse crowd', 'few people present', 'uncrowded area', 'moderate crowd density', 'moderately populated', 'dense crowd', 'tightly packed people', 'overcrowded scene', 'extremely dense crowd' \\
		
		\bottomrule
	\end{tabularx}
	\label{criterion_value}
\end{table*}

\section{Additional Qualitative Comparisons}
Examples of some results are shown in Figure \ref{result_example_appendix}.
\subsection{Case 1}
 In the indoor roller skating scenario, the primary safety hazards revolve around the lack of protective gear and surrounding emergency facilities. The base model recognizes the lack of protective gear but also produces poorly grounded negations, even falsely claiming that no fire extinguishers are visible despite their presence in the scene. The baseline fine-tuned models (SFT, DPO, and ORPO) successfully detect primary omissions, such as the foreground skaters lacking helmets and pads, but fail to identify secondary risks like the absence of a first aid kit on the floor or the unsafe posture of the background skater. GRPO yields an inactive response, asserting that all expected elements are present. In comparison, CRCD reconstructs a latent representation of the expected safe environment (incorporating first aid facilities and fully equipped participants) to contrast with the actual visual features. Consequently, it concurrently identifies the lack of helmets, knee pads, and elbow pads for both skaters, while extracting environmental omissions like the missing first aid kits.

\subsection{Case 2}
This nighttime traffic scenario involves multi-object interactions under low-light conditions. The base model exhibits severe visual grounding errors, falsely negating clearly visible entities such as motorcycles, cars, and people, while generating several irrelevant pedestrian-crossing-related absences. The baseline fine-tuned models exhibit spatial misattributions and perceptual errors. Specifically, SFT and DPO correctly note the missing helmet but attribute this omission to the operator (the woman riding the motorcycle) rather than the child passenger, while also reporting that the vehicle's taillights are inactive. CyclePref similarly claims that neither individual is wearing a helmet. GRPO's reasoning chain focuses on general traffic structures, predicting missing traffic lights and pedestrian crossings that are absent from the local safety context. By employing the Association Branch to map missing entities and anchoring them via the spatial decoder, CRCD localizes the missing helmet to the child passenger and identifies the absence of traffic police officers, avoiding the spatial misattributions observed in the baselines.

\subsection{Case 3}
The interactive tiger cage scenario requires parsing physical barriers and subject identities. The base model fails to ground its negative predictions in the visual scene, repeatedly claiming the absence of visibly present entities such as the person, rather than identifying safety-relevant missing information. The baseline fine-tuned models display class-level confusion and subsequent semantic errors. Specifically, DPO misidentifies the adult woman in the foreground as a little girl, and consequently claims that she lacks protective gear and that no physical barrier exists between her and the predator, overlooking the concrete wall. ORPO inherits this identity mismatch, generating secondary descriptions regarding missing adult guardians and animal muzzles. GRPO fails to generate any negation descriptions. By projecting the scene features onto targeted safety subspaces via the Multi-Condition Representation Learning (MCRL) module, CRCD correctly recognizes the subject's identity as a woman and identifies the lack of restraining ropes or chains on the tiger, alongside the absence of professional handlers and emergency escape routes.

\subsection{Case 4}
In the heavy fog highway scenario, reduced visibility restricts bottom-up visual processing. The base model captures coarse environmental risks such as poor visibility and missing road markings or signs, but its descriptions remain generic and overlook more specific safety-critical omissions, particularly the missing streetlights and inactive vehicle lights. The descriptions from the baseline fine-tuned models (such as DPO and CyclePref) are factually accurate, correctly identifying the poor visibility and the lack of streetlights. However, they are restricted to these immediate environmental factors and fail to identify more specific hazards. By utilizing the Amodal Completion Branch (ACB) to restore degraded object features and incorporating safety priors through the MCRL module, CRCD reconstructs an expected safe road layout in the latent space. As a result, the contrastive decoder identifies more granular semantic differences: in addition to the missing streetlights, it reports that the sedan's headlights are turned off, the road lacks speed limit and warning signs, and there are no guardrails on either side.

\subsection{Case 5}
This industrial scenario involves complex spatial relationships between suspended loads, workers, and site management. The base model partially identifies relevant omissions such as missing safety barriers and warning signs, but also introduces unsupported claims about crane anchoring, lighting, hooks, and outriggers, indicating weak grounding of its negative predictions. The baseline fine-tuned models generate factually correct statements regarding individual safety violations, such as pointing out that the workers are not wearing safety helmets or that the suspended load is unstable. Their analysis, however, remains limited to these individual-level factors. By modeling an expected safe construction environment within the latent space, CRCD expands the scope of negation analysis from isolated objects to system-level hazards. Beyond identifying the missing helmets, the framework's contrastive process detects that the crane operator's view of the pedestrians is obstructed, no safety warning signs are placed around the lifting zone, and there are no safety management personnel on-site to coordinate evacuation.

\label{result_appendix}
\begin{figure*}[!t]
	\centering
	\includegraphics[width=0.95\linewidth]{./pic/result_examples_appendix.jpg}
	\caption{Examples of results under different methods. \textcolor{red}{Red}, \textcolor{blue}{blue}, and \textcolor{green}{green} are used to mark inferred missing objects, attributes, and relationships, respectively.}
	\label{result_example_appendix}
\end{figure*}

}

\end{document}